\documentclass[english]{article}
\pdfoutput=1
\usepackage[T1]{fontenc}
\usepackage[latin9]{inputenc}
\usepackage{geometry}
\usepackage{verbatim}
\usepackage{float}
\usepackage{amsthm}
\usepackage{amsmath}
\usepackage{amssymb}
\usepackage{graphicx}
\usepackage{esint}
\usepackage[numbers]{natbib}

\makeatletter
\numberwithin{equation}{section}
\numberwithin{figure}{section}
\theoremstyle{plain}
\newtheorem{thm}{\protect\theoremname}[section]
  \theoremstyle{definition}
  \newtheorem{defn}[thm]{\protect\definitionname}
  \theoremstyle{definition}
  \newtheorem*{example*}{\protect\examplename}
  \theoremstyle{plain}
  \newtheorem{prop}[thm]{\protect\propositionname}
  \theoremstyle{remark}
  \newtheorem*{rem*}{\protect\remarkname}
  \theoremstyle{plain}
  \newtheorem{cor}[thm]{\protect\corollaryname}
  \theoremstyle{remark}
  \newtheorem*{acknowledgement*}{\protect\acknowledgementname}

\usepackage{times}
\usepackage{tikz}
\usetikzlibrary[topaths]

\@ifundefined{showcaptionsetup}{}{%
 \PassOptionsToPackage{caption=false}{subfig}}
\usepackage{subfig}
\makeatother

\usepackage{babel}
  \providecommand{\acknowledgementname}{Acknowledgement}
  \providecommand{\corollaryname}{Corollary}
  \providecommand{\definitionname}{Definition}
  \providecommand{\examplename}{Example}
  \providecommand{\propositionname}{Proposition}
  \providecommand{\remarkname}{Remark}
\providecommand{\theoremname}{Theorem}

\begin{document}

\title{Automorphism Groups of Graphical Models and Lifted Variational Inference}

\author{Hung H. Bui\\
Artificial Intelligence Center\\
SRI International\\
bui@ai.sri.com\\
\and Tuyen N. Huynh\\
Artificial Intelligence Center\\
SRI International\\
huynh@ai.sri.com\\
\and Sebastian Riedel\\
Department of Computer Science\\
University of Massachusetts, Amherst\\
riedel@cs.umass.edu}

\maketitle
\begin{abstract}
Using the theory of group action, we first introduce the concept of
the \emph{automorphism group} of an exponential family or a graphical
model, thus formalizing the general notion of symmetry of a probabilistic
model. This automorphism group provides a precise mathematical framework
for lifted inference in the general exponential family. Its group
action partitions the set of random variables and feature functions
into equivalent classes (called orbits) having identical marginals
and expectations. Then the inference problem is effectively reduced
to that of computing marginals or expectations for each class, thus
avoiding the need to deal with each individual variable or feature.
We demonstrate the usefulness of this general framework in lifting
two classes of variational approximation for MAP inference: local
LP relaxation and local LP relaxation with cycle constraints; the
latter yields the first lifted inference that operate on a bound tighter
than local constraints. Initial experimental results demonstrate that
lifted MAP inference with cycle constraints achieved the state of
the art performance, obtaining much better objective function values
than local approximation while remaining relatively efficient. \end{abstract}

\global\long\def\Int{\mathbb{N}}

\global\long\def\Real{\mathbb{R}}

\global\long\def\idf#1{\mathbb{I}\left\{  #1\right\}  }

\global\long\def\symg{\mathbb{S}}

\global\long\def\symgn{\symg_{n}}

\global\long\def\symgparti#1{\symg^{c}(#1)}

\global\long\def\Gg{\mathbb{G}}

\global\long\def\intlist#1{\{1,2,\ldots,#1\}}

\global\long\def\subgroup{\le}

\global\long\def\idg{\mathbf{1}}

\global\long\def\aut{\pi}

\global\long\def\Aut{\mathbb{A}}

\global\long\def\toorbit{\rho}

\global\long\def\orbit{\mathrm{orb}}

\global\long\def\Orbit{\mathrm{Orb}}

\global\long\def\stab{\mathrm{Stab}}

\global\long\def\cluster{c}

\global\long\def\Cluster{\mathcal{C}}

\global\long\def\Clustersize#1{\Cluster_{n}^{#1}}

\global\long\def\tup{\tau}

\global\long\def\Tup{\mathcal{T}}

\global\long\def\Tupsize#1{\Tup_{n}^{#1}}

\global\long\def\tupset{\mathrm{set}}

\global\long\def\permvec#1#2{#1_{#2(1)},\ldots,#1_{#2(n)}}

\global\long\def\bp{\mathbf{p}}

\global\long\def\feat{\phi}

\global\long\def\nrfeat{\mathrm{f}}

\global\long\def\Feat{\Phi}

\global\long\def\para{\theta}

\global\long\def\Para{\Theta}

\global\long\def\mean{\mu}

\global\long\def\bmean{\boldsymbol{\mu}}

\global\long\def\Mean{\mathcal{M}}

\global\long\def\outbound{\text{\text{OUT}ER}}

\global\long\def\Local{\text{LOCAL}}

\global\long\def\Cyc{\text{CYCLE}}

\global\long\def\parti{\Delta}

\global\long\def\Index{\mathcal{I}}

\global\long\def\vIndex{\mathcal{V}}

\global\long\def\Ee{\mathcal{E}}

\global\long\def\domain{\mathcal{X}}

\global\long\def\Expt{\mathbb{E}}

\global\long\def\meanmap{\mathbf{m}}

\global\long\def\shatter{\varphi}

\global\long\def\Pot{\Psi}

\global\long\def\pot{\psi}

\global\long\def\naut{\aut}

\global\long\def\paut{\gamma}

\global\long\def\ord{\lambda}

\global\long\def\argl{\mathrm{argl}}

\global\long\def\args{scope}

\global\long\def\argn{\eta}

\global\long\def\argnall#1{|\argn_{#1}|}

\global\long\def\fgraph{\mathcal{F}}

\global\long\def\narg{\kappa}

\global\long\def\cgraph{\mathcal{C}}

\global\long\def\M{\mathcal{M}}
\global\long\def\x{\mathbf{x}}
\global\long\def\E{\text{E}}
\global\long\def\S{\mathcal{S}}
\global\long\def\b{\mathbf{b}}
\global\long\def\R{\mathbb{R}}
\global\long\def\N{\mathbb{N}}

\global\long\def\lmean{\bar{\mean}}

\global\long\def\lparti{\bar{\parti}}

\global\long\def\expandmat{D}

\global\long\def\pAut{\Aut^{p}}

\global\long\def\simparti#1{\stackrel{#1}{\sim}}

\global\long\def\relint{\text{ri\,}}

\global\long\def\riMean{\relint\Mean}

\global\long\def\chull{\text{conv }}

\global\long\def\FF{\mathcal{F}}

\global\long\def\gr{\mathfrak{G}}

\global\long\def\gm{\mathcal{G}}

\global\long\def\gcol{\text{col}}

\global\long\def\formula{F}

\global\long\def\ldom{\mathcal{D}}

\global\long\def\lconsts{\ldom_{0}}

\global\long\def\ldomnovel{\ldom_{*}}

\global\long\def\Ground{\text{Gr}}

\global\long\def\GroundFormula{\text{GrF}}

\global\long\def\model{\omega}

\global\long\def\rename{r}

\global\long\def\onrfeat{\nrfeat^{o}}

\global\long\def\ofeatunary#1#2{\feat_{#1:#2}^{o}}

\global\long\def\ofeatbinary#1#2#3#4{\feat_{\left\{  #1:#3,#2:#4\right\}  }^{o}}

\global\long\def\oMean{\Mean^{o}}

\global\long\def\opara{\para^{o}}

\global\long\def\omean{\mean^{o}}

\global\long\def\oparaunary#1#2{\para_{#1:#2}^{o}}

\global\long\def\oparabinary#1#2#3#4{\para_{\left\{  #1:#3,#2:#4\right\}  }^{o}}

\global\long\def\omeanunary#1#2{\mean_{#1:#2}^{o}}

\global\long\def\omeanbinary#1#2#3#4{\mean_{\left\{  #1:#3,#2:#4\right\}  }^{o}}

\global\long\def\pmeanunary#1#2{\tau_{#1:#2}}

\global\long\def\pmeanbinary#1#2#3#4{\tau_{\left\{  #1:#3,#2:#4\right\}  }}

\global\long\def\liftpmeanunary#1#2{\bar{\tau}_{#1:#2}}

\global\long\def\liftpmeanbinary#1#2#3{\bar{\tau}_{#1:#2#3}}

\global\long\def\oFF{\FF^{o}}

\global\long\def\oIndex{\Index^{o}}

\global\long\def\pair#1#2{#1\text{:}#2}

\section{Introduction}

Classical approaches to probabilistic inference -- an area now reasonably
well understood -- have traditionally exploited low tree-width and
sparsity of the graphical model for efficient exact and approximate
inference. A more recent approach known as \emph{lifted inference
}\citep{braz05lifted,singla08lifted,gogate10exploiting,gogate11probabilistic}
has demonstrated the possibility to perform very efficient inference
in highly-connected, but \emph{symmetric} models such as those arising
in the context of relational (or first-order) probabilistic models.
While it is clear that symmetry is the essential element in lifted
inference, there is currently no formally defined notion of symmetry
of a probabilistic model, and thus no formal account of what ``exploiting
symmetry'' means in lifted inference. 

The mathematical formulation of symmetry of an object is typically
defined via a set of transformations that preserve the object of interest.
Since this set forms a mathematical group (so-called the \emph{automorphism
group} of that object), the theory of groups and group action are
essential in the study of symmetry. 

In this paper, we first introduce the concept of the \emph{automorphism
group} of an exponential family or a graphical model, thus formalizing
the notion of symmetry of a general graphical model. This automorphism
group provides a precise mathematical framework for lifted inference
in graphical models. Its group action partitions the set of random
variables and feature functions into equivalent classes (a.k.a. orbits)
having identical marginals and expectations. The inference problem
is effectively reduced to that of computing marginals or expectations
for each class, thus avoiding the need to deal with each individual
variable or feature. We demonstrate the usefulness of this general
framework in lifting two classes of variational approximation for
MAP inference: local LP relaxation and local LP relaxation with cycle
constraints; the latter yields the first lifted inference that operates
on a bound tighter than local constraints. Initial experimental results
demonstrate that lifted MAP inference with cycle constraints achieved
the state of the art performance, obtaining much better objective
function values than local approximation while remaining relatively
efficient. 

\emph{}%

\section{Background on Groups and Graph Automorphisms}

A \emph{partition} $\parti=\{\parti_{1}\dots\parti_{k}\}$ of a set
$V$ is a set of disjoint nonempty subsets of $V$ whose union is
$V$. Each element $\parti_{i}$ is called a \emph{cell}. A partition
$\parti$ defines an equivalence relation on $V$, denoted as $\simparti{\parti}$,
by letting $u\simparti{\parti}v$ iff $u$ and $v$ are in the same
cell. A partition $\Lambda$ is finer than $\parti$ if every cell
of $\Lambda$ is a subset of some cell of $\parti$. %

We now briefly review some important concepts in group theory and
graph automorphisms \citep{godsil01agt}.

A \emph{group} $(\Gg,\cdot)$ is a non-empty set $\Gg$ with a binary
operation $\cdot$ such that it is associative, closed in $\Gg$;
$\Gg$ contains an identity element, denoted as $\mathbf{1}$, such
that $\forall g\in\Gg,$ $\mathbf{1}\cdot g=g\cdot\mathbf{1}=g$ and
there exists an element $g^{-1}$ such that $g\cdot g^{-1}=g^{-1}\cdot g=\mathbf{1}$.
A group containing $\idg$ as its only element is called a trivial\emph{
group. A} \emph{subgroup} of $\Gg$ is a subset of $\Gg$ that forms
a group with the same binary operation as $\Gg$. We write $\Gg_{1}\subgroup\Gg_{2}$
when $\Gg_{1}$ is a subgroup%
\footnote{We use the notation $\Gg_{1}\preceq\Gg_{2}$ to mean $\Gg_{1}$ is
isomorphic to a subgroup of $\Gg_{2}$.%
} of $\Gg_{2}$.

A permutation on a set $V$ is a bijective mapping from $V$ to itself.
The set of all permutations of $V$ together with the mapping-composition
operator forms a group named the \emph{symmetric group} $\symg(V)$.
A symmetric group that plays a central role in this paper is the symmetric
group\emph{ $\symg_{n}$}, the set of all permutations of $\intlist n$.
For a permutation $\pi\in\symg_{n}$, $\pi(i)$ is the image of $i$
under $\pi$. For each vector $x\in\mathcal{X}^{n}$, the vector $x$
permuted by $\pi$, denoted by $x^{\pi}$, is $(x_{\pi(1)}\ldots x_{\pi(n)})$;
for a set $A\subset\mathcal{X}^{n}$, the set $A$ permuted by $\pi$,
denoted by $A^{\pi}$ is $\left\{ x^{\pi}|x\in A\right\} $.

The \emph{action} of a group $\Gg$ \emph{on a set} $V$ is a mapping
that assigns every $g\in\Gg$ to a permutation on $V$, denoted as
$g():V\rightarrow V$ such that the identity element $\mathbf{1}$
is assigned to the identity permutation, and the group product of
two elements $g_{1}\cdot g_{2}$ is assigned to the composition $g_{1}()\circ g_{2}()$.
The action of a group $\Gg$ on $V$ induces an equivalence relation
on $V$ defined as $v\sim v'$ iff there exists $g\in\Gg$ such that
$g(v)=v'$ (the fact that $\sim$ is an equivalence relation follows
from the definition of group). The group action therefore induces
a partition on $V$ called the \emph{orbit partition,} denoted as
$\Orbit_{\Gg}(V)$. The \emph{orbit} of an element $v\in V$ under
the action of $\Gg$ is the set of elements in $V$ equivalent to
v: $\orbit_{\Gg}(v)=\{v^{\prime}\in\vIndex|\: v^{\prime}\sim v\}$.
Any subgroup $\Gg_{1}\subgroup\Gg$ will also act on $V$ and induces
a finer equivalence relation (and hence a more refined orbit partition).
Given $v\in V$, if under the group action, every element $g\in\Gg$
preserves $v$, that is $\forall g\in\Gg$, g(v)=v, then the group
$\Gg$ is said to stabilize $v$. %
{} %

Next, we consider the action of a permutation group on the vertex
set of graph, which leads to the concept of graph automorphisms. 

An \emph{automorphism} of a graph $\gr$ on a set of vertices $V$
is a permutation $\pi\in\symg(V)$ that permutes the vertices of $\gr$
but preserves the structure (e.g., adjacency, direction, color) of
$\gr.$%
{} The set of all automorphisms of $\gr$ forms a group named the \emph{automorphism
group} of $\gr$, denoted as $\Aut(\gr)$. It is clear that $\Aut(\gr)$
is a subgroup of $\symg(V)$. The cardinality of $\Aut(\gr)$ indicates
the level of symmetry in $\gr$; if $\Aut(\gr)$ is the trivial group
then $\gr$ is asymmetric.

The action of $\Aut(\gr)$ on the vertex set $V$ partitions $V$
into the node-orbits $\Orbit_{\Aut(\gr)}(V)$ where each node orbit
is a set of vertices equivalent to one another up to some node relabeling.
Furthermore, $\Aut(\gr)$ also acts on the set of graph edges $E$
by letting $\pi(\{u,v\})=\{\pi(u),\pi(v)\}$ and this action partitions
$E$ into a set of edge-orbits $\Orbit_{\Aut(\gr)}(E)$. Similarly,
we also obtain the set of arc-orbits $\Orbit_{\Aut(\gr)}(\stackrel{\rightarrow}{E})$.

Computing the automorphism group of a graph is as difficult as determining
whether two graphs are isomorphic, a problem that is known to be in
NP, but for which it is unknown whether it has a polynomial time algorithm
or is NP-complete. In practice, there exists efficient computer programs
such as \emph{nauty}%
\footnote{\emph{http://cs.anu.edu.au/people/bdm/nauty/}%
} \citep{mckay81pgi} for computing automorphism groups of graphs.

\section{Symmetry of the Exponential Family}

\subsection{Exponential Family and Graphical Model}

Consider an exponential family over $n$ random variables $(x_{i})_{i\in\mathcal{V}}$
where $\vIndex=\left\{ 1\ldots n\right\} $, $x_{i}\in\domain$ with
density function
\[
\FF(x\,|\,\theta)=h(x)\exp\left(\left\langle \Feat(x),\theta\right\rangle -A(\theta\right))
\]
 where $h$ is the base density, $\Feat(x)=(\feat_{j}(x))_{j\in\Index}$,
$\Index=\intlist m$ is an $m$-dimensional feature vector, $\theta\in\Real^{m}$
is the natural parameter, and $A(\theta)$ the log-partition function.
Let $\Para=\{\theta\,|A(\theta)<\infty\}$ be the set of natural parameters,
$\Mean=\left\{ \mean\in\Real^{m}\ |\ \exists p,\:\mean=\E_{p}\Feat(x)\right\} $
the set of realizable mean parameters, $A^{*}:\Mean\rightarrow\Real$
the convex dual of $A$, and $\meanmap:\Para\rightarrow\Mean$ the
mean parameter mapping that maps $\para\mapsto\meanmap(\para)=\E_{\para}\Feat(x)$.
Note that $\meanmap(\Para)=\riMean$ is the relative interior of $\Mean$.
For more details, see \citep{wainwright2008graphical}.

Often, a feature function $\feat_{i}$ depends only on a subset of
the variables in $\vIndex$. In this case we will write $\feat_{i}$
more compactly in factorized form as $\feat_{i}(x)=\nrfeat_{i}(x_{i_{1}}\ldots x_{i_{K}})$
where the indices $i_{j}$ are distinct, $i_{1}<i_{2}\ldots<i_{K}$,
and $\nrfeat_{i}$ cannot be reduced further, i.e., it must depend
on all of its arguments. To keep track of variable indices of arguments
of $\nrfeat_{i}$, we let $\args(\nrfeat_{i})$ denote its set of
arguments, $\argn_{i}(k)=i_{k}$ the $k$-th argument and $\argnall i$
its number of arguments. Factored forms of features can be encoded
as a hypergraph $\gm\left[\FF\right]$ of $\FF$ (called the graph
structure or graphical model of $\FF$) with nodes $\vIndex$, and
hyperedges (clusters) $\left\{ C|\exists i,\args(\nrfeat_{i})=C\right\} $.
For models with pairwise features, $\gm$ is a standard graph.

For discrete random variables (i.e., $\domain$ is finite), we often
want to work with the overcomplete family $\oFF$ that we now describe
for the case with pairwise features. The set of overcomplete features
$\oIndex$ are indicator functions on the nodes and edges of the graphical
model $\gm$ of $\FF$: $\ofeatunary ut(x)=\idf{x_{u}=t},t\in\domain$
for each node $u\in V(\gm)$; and $\ofeatbinary uvt{t'}(x)=\idf{x_{u}=t,x_{v}=t'},t,t'\in\domain$
for each edge $\{u,v\}\in E(\gm)$. The set of overcomplete realizable
mean parameters $\oMean$ is also called the \emph{marginal polytope
}since the overcomplete mean parameter corresponds to node and edge
marginal probabilities.\emph{ }Given a parameter $\para$, the transformation
of $\FF(x|\para)$ to its overcomplete representation is done by letting
$\para^{o}$ be the corresponding parameter in the overcomplete family:
$\oparaunary ut=\sum_{i\,\text{s.t. }\args(\nrfeat_{i})=\left\{ u\right\} }\nrfeat_{i}(t)\para_{i}$
and (assuming $u<v$) $\oparabinary uvt{t'}=\sum_{i\,\text{s.t. }\args(\nrfeat_{i})=\left\{ u,v\right\} }\nrfeat_{i}(t,t')\para_{i}$.
It is straightforward to verify that $\FF^{o}(x|\para^{o})=\FF(x|\para)$.

\subsection{\label{subsec:aut-exp-family}Automorphism Group of an Exponential
Family}

We define the symmetry of an exponential family $\FF$ as the group
of transformations that preserve $\FF$ (hence preserve $h$ and $\Feat$).
The kind of transformation used will be a pair of permutations $(\naut,\paut)$
where $\naut$ permutes the set of variables and $\paut$ permutes
the feature vector. 
\begin{defn}
An automorphism of the exponential family $F$ is a pair of permutations
$(\naut,\paut)$ where $\naut\in\symg_{n}$, $\paut\in\symg_{m}$
such that for all vectors $x$: $h(x^{\naut})=h(x)$ and $\Feat^{\paut^{-1}}(x^{\naut})=\Feat(x)$
(or equivalently, $\Feat(x^{\naut})=\Feat^{\paut}(x)$).

It is straightforward to show that the set of all automorphisms of
$\FF$, denoted by $\Aut[\FF]$, forms a subgroup of $\symg_{n}\times\symg_{m}$.
This group acts on $\Index$ by the permuting action of $\paut$,
and on $\vIndex$ by the permuting action of $\naut$. In the remainder
of this paper, $h$ is always a symmetric function (e.g., $h\equiv1$);
therefore, the condition $h(x^{\naut})=h(x)$ automatically holds. \end{defn}
\begin{example*}
Let $\vIndex=\{1,2,3\}$ and $\Phi=\left\{ \nrfeat_{1},\nrfeat_{2},\nrfeat_{3}\right\} $
where $\nrfeat_{1}(x_{1},x_{2})=x_{1}(1-x_{2})$, $\nrfeat_{2}(x_{1},x_{3})=x_{1}(1-x_{3})$,
and $\nrfeat_{3}(x_{2},x_{3})=x_{2}x_{3}$. The pair of permutations
$(\naut,\paut)$ where $\pi=(1\mapsto1,\,2\mapsto3,\,3\mapsto2)$
and $\gamma=(1\mapsto2,\,2\mapsto1,\,3\mapsto3)$ is an automorphism
of $\FF$, since $\Phi^{\gamma^{-1}}(x^{\naut})=(\feat_{2}(x_{1},x_{3},x_{2}),\feat_{1}(x_{1},x_{3},x_{2}),\feat_{3}(x_{1},x_{3},x_{2}))=(\nrfeat_{2}(x_{1},x_{2}),\nrfeat_{1}(x_{1},x_{3}),\nrfeat_{3}(x_{3},x_{2}))=(x_{1}(1-x_{2}),x_{1}(1-x_{3}),x_{3}x_{2})=\Phi(x_{1},x_{2,}x_{3})$.

\end{example*}
An automorphism $(\naut,\paut)$ can be characterized in terms of
the factorized features $\nrfeat_{i}$ as follow.
\begin{prop}
\label{prop:aut-characterization}$(\naut,\paut)$ is an automorphism
of $\FF$ if and only if the following conditions are true for all
$i\in\Index$: (1) $\argnall i=\argnall{\paut(i)}$; (2) $\naut$
is a bijective mapping from $\args(\nrfeat_{i})$ to $\args(\nrfeat_{\paut(i)})$;
(3) let $\alpha=\argn_{\paut(i)}^{-1}\circ\naut\circ\argn_{i}$ then
$\alpha\in\symg_{\argnall i}$ and $\nrfeat_{i}(t^{\alpha})=\nrfeat_{\paut(i)}(t)$
for all $t$$\in\domain^{\argnall i}$.\end{prop}
\begin{rem*}
Consider automorphisms of the type $(\idg,\paut)$: $\paut$ must
permute between the features having the same scope: $\args(\nrfeat_{i})=\args(\nrfeat_{\paut(i)})$.
Thus if the features do not have redundant scopes (i.e., $\args(\nrfeat_{i})\neq\args(\nrfeat_{j})$
when $i\neq j$) then $\paut$ must be $\idg$. More generally when
features do not have redundant scopes, $\naut$ uniquely determines
$\paut$. Next, consider automorphisms of the type $(\naut,\idg)$:
$\naut$ must permute among variables in a way that preserve all the
features $\nrfeat_{i}$. Thus if all features are asymmetric functions
then $\naut$ must be $\idg$; more generally, $\paut$ uniquely determines
$\naut$. As a consequence, if the features do not have redundant
scopes and are asymmetric functions then there exists a one-to-one
correspondence between $\naut$ and $\paut$ that form an automorphism
in $\Aut[\FF]$.
\end{rem*}

An automorphism defined above preserves a number of key characteristics
of the exponential family $\FF$ (such as its natural parameter space,
its mean parameter space, its log-partition function), as shown in
the following theorem.
\begin{thm}
\label{theorem:aut-properties}If $(\naut,\paut)\in\Aut[\FF]$ then \end{thm}
\begin{enumerate}
\item $\naut\in\Aut(\gm[\FF])$, i.e. $\naut$ is an automorphism of the
graphical model graph $\gm[\FF]$.
\item $\Para^{\paut}=\Para$ and $A(\para^{\paut})=A(\para)$ for all $\theta\in\Para$. 
\item $\FF(x^{\naut}|\para^{\paut})=\FF(x|\para)$ for all $x\in\domain^{n}$,
$\theta\in\Para$.
\item $\meanmap^{\paut}(\para)=\meanmap(\para^{\paut})$ for all $\theta\in\Para$.
\item $\Mean^{\paut}=\Mean$ and $A^{*}(\mean^{\paut})=A^{*}(\mean)$ for
all $\mean\in\Mean$.\end{enumerate}

\section{Lifted Variational Inference Framework}

We now discuss the principle of how to exploit the symmetry of the
exponential family graphical model for lifted variational inference.
In the general variational inference framework~\citep{wainwright2008graphical},
marginal inference is viewed as to compute the mean parameter $\mean=\meanmap(\para)$
given a natural parameter $\para$ by solving the optimization problem
\begin{equation}
\sup_{\mean\in\Mean}\left\langle \para,\mean\right\rangle -A^{*}(\mean).\label{eq:variational-mean}
\end{equation}
For discrete models, the variational problem is more conveniently
posed using the overcomplete parameterization, for marginal inference
\begin{equation}
\sup_{\omean\in\oMean}\left\langle \omean,\opara\right\rangle -A{}^{o*}(\omean)\label{eq:variational-marginal}
\end{equation}
 and for MAP inference
\begin{equation}
\max_{x\in\domain^{n}}\ln\FF(x|\para)=\sup_{\omean\in\oMean}\left\langle \omean,\opara\right\rangle +\text{const}.\label{eq:variational-map}
\end{equation}

We first focus on lifting the main variational problem in (\ref{eq:variational-mean})
and leave discussions of the other problems to subsection \ref{subsec:overcomplete}.

\subsection{Parameter Tying and Lifting Partition}

Lifted inference in essence assumes a parameter-tying setting where
some components of $\para$ are the same. More precisely, we assume
a partition $\parti$ of $\Index$ (called the \emph{parameter-tying
partition}) such that $j\simparti{\parti}j^{'}\Rightarrow\para_{j}=\para_{j^{'}}$.
Our goal is to study how parameter-tying, coupled with the symmetry
of the family $\FF$, can lead to more efficient variational inference.

Let $\Real_{\parti}^{m}$ denote the subspace $\left\{ r\in\Real^{m}\,|\, r_{j}=r_{j^{'}}\,\text{if }j\stackrel{\parti}{\sim}j^{'}\right\} $.
For any set $S\subset\Real^{m}$, let $S_{\parti}=S\cap\Real_{\parti}^{m}$.
Restricting the natural parameter to $\Para_{\parti}$ is equivalent
to parameter tying, and hence, equivalent to working with a different
exponential family with $|\parti|$ aggregating features $\left(\sum_{j\in\parti_{i}}\feat_{j}\right)_{i}$.
While this family has fewer parameters, it is not obvious how it would
help inference; moreover, in working directly with the aggregation
features, the structure of the original family is lost. 

To investigate the effect parameter tying has on the complexity of
inference, we turn to the question of how to characterize the image
of $\Para_{\parti}$ under the mean mapping $\meanmap$. At first,
note that in general $\meanmap(\Para_{\parti})\neq\Mean_{\parti}$:
taking $\parti$ to be the singleton partition $\left\{ \Index\right\} $
will enforce all natural parameters to be the same, but clearly this
does not guarantee that all mean parameters are the same. However,
one can hope that perhaps some mean parameters are forced to be the
same due to the symmetry of the graphical model. More precisely, we
ask the following question: is there a partition $\shatter$ of $\Index$
such that for all $\para\in\Para_{\parti}$ the mean parameter is
guaranteed to lie inside $\Mean_{\shatter}$, and therefore the domain
of the variational problem (\ref{eq:variational-mean}) can be restricted
accordingly to $\Mean_{\shatter}$? Such partitions are defined for
general convex optimization problems below.
\begin{defn}
(Lifting partition) Consider the convex optimization $inf{}_{\mathbf{x}\in\mathcal{S}}\, J(\mathbf{x})$
where $\mathcal{S}\subset\Real^{m}$ is a convex set and $J$ is a
convex function. A partition $\varphi$ of $\left\{ 1\ldots m\right\} $
is a \emph{lifting partition} for the aforementioned problem iff $inf{}_{x\in S}\, J(x)\:=inf_{x\in S_{\shatter}}J(x)$,
i.e., the constraint set $S$ can be restricted to $S_{\shatter}$.\end{defn}
\begin{thm}
\label{theorem:lifting-convex-optimization}Let $\Gg$ act on $I=\left\{ 1\ldots m\right\} $,
so that every $g\in\Gg$ corresponds to some permutation on $\left\{ 1\ldots m\right\} $.
If $S^{g}=S$ and $J(x^{g})=J(x)$ for every $g\in\Gg$ (i.e., $\Gg$
stabilizes both $S$ and $J$) then the induced orbit partition $\Orbit_{\Gg}(I)$
is a lifting partition for $inf{}_{x\in\mathcal{S}}\, J(x)$.
\end{thm}
From theorem \ref{theorem:aut-properties}, we know that $\Aut[\FF]$
stabilizes $\Mean$ and $A^{*}$; however, this group does not take
the parameter $\para$ into account. Given a partition $\parti$,
a permutation $\lambda$ on $\Index$ is consistent with $\parti$
iff $\lambda$ permutes only among elements of the same cell of $\parti$.
Such permutations are of special interest since for every $\para\in\Para_{\parti}$,
$\para^{\lambda}=\para$. If $\Gg$ is a group acting on $\Index$,
we denote $\Gg_{\parti}$ the set of group elements whose actions
are consistent with $\parti$, that is $\Gg_{\parti}=\left\{ g\in\Gg|\forall u\in\Index,\, g(u)\simparti{\parti}u\right\} $.
It is straightforward to verify that $\Gg_{\parti}$ is a subgroup
of $\Gg$. With this notation, $\Aut_{\parti}(\FF)$ is the subgroup
of $\Aut[\FF]$ whose member's action is consistent with $\parti$.
The group $\Aut_{\parti}(\FF)$ thus stabilizes not just the family
$\FF$, but also every parameter $\para\in\Para_{\parti}$. It is
straightforward to verify $\Aut_{\parti}(\FF)$ stabilizes both the
constraint set and the objective function of (\ref{eq:variational-mean}).
Therefore by the previous theorem, its induced orbit yields a lifting
partition. %

\begin{cor}
\label{cor:lifting-partition-mean} Let $\shatter=\shatter(\parti)=\Orbit_{\Aut_{\parti}[\FF]}(\Index)$.
Then for all $\para\in\Para_{\parti}$, $\shatter$ is a lifting partition
for the variational problem (\ref{eq:variational-mean}), that is
\begin{equation}
\sup_{\mean\in\Mean}\left\langle \para,\mean\right\rangle -A^{*}(\mean)=\sup_{\mean\in\Mean_{\shatter}}\left\langle \para,\mean\right\rangle -A^{*}(\mean)\label{eq:lifted-mean}
\end{equation}

\end{cor}
In (\ref{eq:lifted-mean}), we call the LHS the \emph{ground} formulation
of the variational problem, and the RHS the \emph{lifted} formulation.
Let $\ell=|\shatter|$ be the number of cells of $\shatter$, the
\emph{lifted} constraint set $\M_{\shatter}$then effectively lies
inside an $\ell$-dimensional subspace where $\ell\le m$. This forms
the core idea of the principle of lifted variational inference: to
perform optimization over the lower dimensional (and hopefully easier)
constraint set $\M_{\shatter}$ instead of $\M$. 
\begin{rem*}
The above result also holds for \emph{any subgroup} $\Gg$ of $\Aut_{\parti}(\FF)$
since $\shatter_{\Gg}=\Orbit_{\Gg}(\Index)$ is finer than $\shatter$.
Thus, it is obvious that $\shatter_{\Gg}$ is also a lifting partition.
However, the smaller is the group $\Gg$, the finer is the lifting
partition $\shatter_{\Gg}$, and the less symmetry can be exploited.
In the extreme, $\Gg$ can be the trivial group, $\shatter_{\Gg}$
is the discrete partition on $\Index$ putting each element in its
own cell, and $\Mean_{\shatter_{\Gg}}=\Mean$, which corresponds to
no lifting.
\end{rem*}

\textbf{}%

\subsection{Characterization of $\Mean_{\shatter}$}

\label{sec:lifted_procedure}

We now give a characterization of $\Mean_{\shatter}$ in the case
of discrete random variables. Note that $\Mean$ is the convex hull
$\Mean=\chull\left\{ \Feat(x)|x\in\domain^{n}\right\} $ which is
a polytope in $\Real^{m}$, and $\Aut[\FF]$ acts on the set of configurations
$\domain^{n}$ by the permuting action of $\naut$ which maps $x\mapsto x^{\naut}$. 
\begin{thm}
\label{theorem:lifted-extremes}Let $\mathcal{O}=\Orbit_{\Aut_{\parti}[\FF]}(\domain^{n})$
be the set of $\mathcal{X}$-configuration orbits. For each orbit
$\mathcal{C}\in\mathcal{O}$, let $\bar{\Feat}(\mathcal{C})=\frac{1}{|\mathcal{C}|}\sum_{x\in\mathcal{C}}\Phi(x)$
be the feature-centroid of all the configurations $x$ in $\mathcal{C}$.
Then $\Mean_{\shatter(\parti)}=\chull\left\{ \bar{\Phi}(\mathcal{C})|\mathcal{C}\in\mathcal{O}\right\} $. 
\end{thm}
As a consequence, the \emph{lifted polytope} $\Mean_{\shatter}$ can
have at most $|\mathcal{O}|$ extreme points. The number of configuration
orbits $|\mathcal{O}|$ can be much smaller than the total number
of configurations $|\domain|^{n}$ when the model is highly symmetric.
For example, for a fully connected graphical model with identical
pairwise and unary potentials and $\domain=\left\{ 0,1\right\} $
then every permutation $\naut\in\symg_{n}$ is part of an automorphism;
thus, every configuration with the same number of $1$'s belongs to
the same orbit, and hence $|\mathcal{O}|=n+1$. In general, however,
$|\mathcal{O}|$ often is still exponential in $n$. We discuss approximations
of $\Mean_{\shatter}$ in Section \ref{sec:lifted-approximate-map}. 

A representation of the lifted polytope $\Mean_{\shatter}$ by a set
of constraints in $\Real^{|\shatter|}$ can be directly obtained from
the constraints of the polytope $\Mean$. For each cell $\shatter_{j}$
($j=1,\ldots,|\shatter|)$ of $\shatter$, let $\bar{\mean}_{j}$
be the common value of the variables $\mean_{i}$, $i\in\shatter_{j}$.
Let $\rho$ be \emph{the orbit mapping function} that maps each element
$i\in\Index$ to the corresponding cell $\rho(i)=j$ that contains
$i$. Substituting $\mean_{i}$ by $\bar{\mean}_{\rho(i)}$ in the
constraints of $\Mean$, we obtain a set of constraints in $\bar{\mean}$
(in vector form, we substitute $\mean$ by $\expandmat\bar{\mean}$
where $\expandmat_{ij}=1$ if $i\in\shatter_{j}$ and $0$ otherwise).
In doing this, some constraints will become identical and thus redundant.
In general, the number of non-redundant constraints can still be exponential.

\subsection{\label{subsec:overcomplete}Overcomplete Variational Problems}

We now state analogous results in lifting the overcomplete variational
problems (\ref{eq:variational-marginal}) and (\ref{eq:variational-map})
when $\domain$ is finite. To simplify notation, we will consider
only the case where features are unary or pairwise. As before, the
group $\Aut_{\parti}[\FF]$ will be used to induce a lifting partition.
However, we need to define the action of this group on the set of
overcomplete features $\oIndex$. 

Recall that if $(\naut,\paut)\in\Aut[\FF]$ then $\naut$ is an automorphism
of the graphical model graph $\gm$. Since overcomplete features naturally
correspond to nodes and edges of $\gm$, $\naut$ has a natural action
on $\oIndex$ that maps $\pair vt\mapsto\pair{\naut(v)}t$ and $\left\{ \pair ut,\pair v{t'}\right\} \mapsto\left\{ \pair{\naut(u)}t,\pair{\naut(v)}{t'}\right\} $.
Define $\shatter^{o}=\shatter^{o}(\parti)=\Orbit_{\Aut_{\parti}[\FF]}(\oIndex)$
to be the induced orbits of $\Aut_{\parti}[\FF]$ on the set of overcomplete
features. 
\begin{cor}
\label{cor:lifted-variational-overcomplete}For all $\para\in\Para_{\parti}$,
$\shatter^{o}$ is a lifting partition for the variational problems
(\ref{eq:variational-marginal}) and (\ref{eq:variational-map}).
\end{cor}
Thus, the optimization domain can be restricted to $\Mean_{\shatter^{o}}^{o}$
which we term the \emph{lifted marginal polytope}. The cells of $\shatter^{o}$
are intimately connected to the node, edge and arc orbits of the graph
$\gm$ induced by $\Aut_{\parti}[\FF]$. We now list all the cells
of $\shatter^{o}$ in the case where $\domain=\left\{ 0,1\right\} $:
each node orbit $\bar{v}$ corresponds to 2 cells $\left\{ v:t|v\in\bar{v}\right\} ,t\in\left\{ 0,1\right\} $;
each edge orbit $\bar{e}$ corresponds to 2 cells $\left\{ \left\{ u:t,v:t\right\} |\left\{ u,v\right\} \in\bar{e}\right\} ,t\in\left\{ 0,1\right\} $;
and each arc orbit $\bar{a}$ corresponds to the cell $\left\{ \left\{ u:0,v:1\right\} |(u,v)\in\bar{a}\right\} $.
The orbit mapping function $\rho$ maps each element of $\oIndex$
to its orbit as follows: $\rho(\pair vt)=\pair{\bar{v}}t$, $\rho(\{\pair ut,\pair vt\})=\pair{\{\overline{u,v}\}}t$,
$\rho(\{\pair u0,\pair v1\})=\pair{(\overline{u,v})}{01}$. \textbf{}%

The total number of cells of $\shatter^{o}$ is $O(|\bar{V}|+|\bar{E}|)$
where $|\bar{V}|$ and $|\bar{E}|$ are the number of node and edge
orbits of $\gm$ (each edge orbit corresponds to at most 2 arc orbits).
Thus, in working with $\Mean_{\shatter^{o}}^{o}$, the big-$O$ order
of the number of variables is reduced from the number of nodes and
edges in $\gm$ to the number of node and edge orbits.%

\section{\label{sec:lifted-approximate-map}Lifted Approximate MAP Inference}

Approximate variational inference typically works with a tractable
approximation of $\Mean$ and a tractable approximation of $A^{*}$.
In this paper, we focus only on lifted outer bounds of $\Mean^{o}$
(and thus restrict ourselves to the discrete case). We leave the problem
of handling approximations of $A^{*}$ to future work. Thus, our focus
will be on the LP relaxation of the MAP inference problem (\ref{eq:variational-map}). 

By corollary \ref{cor:lifted-variational-overcomplete}, (\ref{eq:variational-map})
is equivalent to the lifted problem $\sup_{\omean\in\Mean_{\shatter^{o}}^{o}}\left\langle \opara,\omean\right\rangle $.
Since any outer bound $\outbound\supset\oMean$ yields an outer bound
$\outbound_{\shatter^{o}}$ of $\Mean_{\shatter^{o}}^{o}$, we can
always relax the lifted problem and replace $\Mean_{\shatter^{o}}$
by $\outbound_{\shatter^{o}}$. But is the relaxed lifted problem
on $\outbound_{\shatter^{o}}$ equivalent to the relaxed ground problem
on $\outbound$? This depends on whether $\shatter^{o}$ is a lifting
partition for the relaxed ground problem.
\begin{thm}
\label{theorem:lifted-outer}If the set $\outbound=\outbound(\gm)$
depends only on the graphical model structure $\gm$ of $\FF$, then
for all $\para\in\Para_{\parti}$, $\shatter^{o}$ is a lifting partition
for the relaxed MAP problem 
\[
\sup_{\omean\in\outbound}\left\langle \opara,\omean\right\rangle =\sup_{\omean\in\outbound_{\shatter^{o}}}\left\langle \opara,\omean\right\rangle 
\]

\end{thm}
The most often used outer bound of $\Mean^{o}$ is the local marginal
polytope $\Local(\gm)$~\citep{wainwright2008graphical}, which enforces
consistency for marginals on nodes and between nodes and edges of
$\gm$. \citep{sontag07newouter,sontag10thesis} used $\Cyc(\gm)$,
which is a tighter bound that also enforces consistency of edge marginals
on the same cycle of $\gm$\emph{. }The Sherali-Adams hierarchy%
\footnote{A note about terminology: Following the tradition in lifted inference,
this paper uses the term \emph{lift} to refer to the exploitation
of symmetry for avoiding doing inference on the \emph{ground} model.
It is unfortunate that the term \emph{lift} has also been used in
the context of coming up with better bounds for the marginal polytopes.
There, \emph{lift} (as in lift-and-project) means to move to a higher
dimensional space where constraints can be more easily expressed with
auxiliary variables.%
} \citep{sherali90hierarchy} provides a sequence of outer bounds of
$\Mean^{o}$, starting from $\Local(\gm)$ and progressively tightening
it to the exact marginal polytope $\Mean^{o}$. All of these outer
bounds depend only on the structure of the graphical model $\gm$,
and thus the corresponding relaxed MAP problems admit $\shatter^{o}$
as a lifting partition. Note that with the exception when $\outbound=\Local$,
equitable partitions~\citep{godsil01agt} of $\gm$ such as those
used in \citep{mladenov12aistats} are not lifting partitions for
the approximate variational problem in theorem \ref{theorem:lifted-outer}.%
\footnote{As a counter example, consider a graphical model whose structure is
the Frucht graph (http://en.wikipedia.org/wiki/Frucht\_graph). Since
this is a regular graph, LOCAL approximation yields identical constraints
for every node. However, the nodes on this graph participate in cycles
of different length, hence are subject to different cycle constraints.%
}

\global\long\def\llparti{\bar{\Delta}_{L}}

\global\long\def\ltau{\bar{\tau}}

\global\long\def\lLocal{\Local_{\llparti}}

\global\long\def\ubar{\bar{u}}

\global\long\def\vbar{\bar{v}}

\global\long\def\ebar{\bar{e}}

\global\long\def\abar{\bar{a}}

\global\long\def\vbold{\mathbf{v}}

\global\long\def\ubold{\mathbf{u}}

\global\long\def\ebold{\mathbf{e}}

\global\long\def\abold{\mathbf{a}}

\global\long\def\taubar{\bar{\tau}}

\global\long\def\thetabar{\bar{\theta}}

\global\long\def\Vbar{\bar{V}}

\global\long\def\Ebar{\bar{E}}

\global\long\def\Gbar{\bar{G}}

\global\long\def\pmean{\tau}

\global\long\def\pmeanlift{\bar{\pmean}}

\section{Lifted MAP Inference on the Local Polytope}

We now focus on lifted approximate MAP inference using the local marginal
polytope $\Local$. From this point on, we also restrict ourselves
to models where the features are pairwise or unary, and variables
are binary ($\domain=\left\{ 0,1\right\} $). 

We first aim to give an explicit characterization of the constraints
of the lifted local polytope $\Local_{\shatter^{o}}$. The local polytope
$\Local(\gm)$ is defined as the set of locally consistent pseudo-marginals.
{\small 
\[
\left\{ \pmean\geq0\left|\begin{array}{cc}
\pmean_{v:0}+\pmean_{v:1}=1 & \forall v\in\vIndex(\gm)\\
\pmean_{\left\{ u:0,v:0\right\} }+\pmean_{\{u:0,v:1\}}=\pmean_{u:0}\\
\pmean_{\{u:0,v:0\}}+\pmean_{\{v:0,u:1\}}=\pmean_{v:0} & \forall\left\{ u,v\right\} \in E(\gm)\\
\tau_{\left\{ u:1,v:1\right\} }+\tau_{\{u:0,v:1\}}=\tau_{v:1}\\
\pmean_{\{u:1,v:1\}}+\pmean_{\{v:0,u:1\}}=\pmean_{u:1}
\end{array}\right.\right\} 
\]
}{\small \par}

Substituting $\tau_{i}$ by the corresponding $\taubar_{\rho(i)}$
where $\rho()$ is given in subsection \ref{subsec:overcomplete},
and by noting that constraints generated by $\{u,v\}$ in the same
edge orbits are redundant, we obtain the constraints for the \emph{lifted
local polytope} $\Local_{\shatter^{o}}${\small{} as follows.
\[
\left\{ \pmeanlift\geq0\left|\begin{array}{cc}
\pmeanlift_{\vbar:0}+\pmeanlift_{\vbar:1}=1 & \forall\text{ node orbit }\vbar\\
\pmeanlift_{\bar{e}:00}+\pmeanlift_{(\overline{u,v}):01}=\pmeanlift_{\ubar:0}\\
\pmeanlift_{\bar{e}:00}+\pmeanlift_{(\overline{v,u}):01}=\pmeanlift_{\vbar:0} & \forall\text{ edge orbit }\bar{e}\\
\taubar_{\bar{e}:11}+\taubar_{(\overline{u,v}):01}=\taubar_{\vbar:1} & (\overline{u,v}),(\overline{v,u}):\text{arc}\\
\taubar_{\bar{e}:11}+\taubar_{(\overline{v,u}):01}=\taubar_{\ubar:1} & \text{orbits of}\bar{\; e}
\end{array}\right.\right\} 
\]
}%
{} Thus, the number of constraints needed to describe the lifted local
polytope $\Local_{\shatter^{o}}$ is $O(|\Vbar|+|\Ebar|)$. Similar
to the ground problem, these constraints can be derived from a graph
representation of the node and edge orbits. Define the \emph{lifted
graph} $\bar{\gm}$ be a graph whose nodes are the set of node orbits
$\bar{V}$ of $\gm$. For each edge orbit $\bar{e}$ with a representative
$\left\{ u,v\right\} \in\bar{e}$, there is a corresponding edge on
$\bar{\gm}$ that connects the two node orbits $\ubar$ and $\vbar$.
Note that unlike $\gm$, the lifted graph $\bar{\gm}$ in general
is not a simple graph and can contain self-loops and multi-edges between
two nodes. Figure 6.1 shows the ground graph $\gm$ and the lifted
graph $\bar{\gm}$ for the example described in subsection \ref{subsec:aut-exp-family}.%

\begin{figure}[H]
\begin{center}
\tikzstyle{var}=[circle,draw=blue!50,fill=blue!20,thick] 
\begin{tikzpicture}
\node (var1) at ( 2,1.2) [var] {1}; 
\node (var2) at ( 1,0) [var] {2}; 
\node (var3) at ( 3,0) [var] {3}; 
\draw (var2) -- (var1); 
\draw (var3) -- (var1); 
\draw (var2) -- (var3);
\pgftext [base,x=2cm,y=-1cm]{a. Ground graph $\mathcal{G}$}
\node (var4) at (7,1.2) [var] {1}; 
\node (var5) at (7,0) [var] {2,3}; 
\draw (var4) -- (var5); 
\draw (var5) to [loop right] ();
\pgftext[base,x=7cm,y=-1cm] {b. Lifted graph $\bar{\mathcal{G}}$}
\end{tikzpicture} 
\end{center}

\caption{$\gm$ and $\bar{\gm}$ of the example described in section \ref{subsec:aut-exp-family}}
\end{figure}

We now consider the linear objective function $\left\langle \opara,\tau\right\rangle $.
Substituting $\tau_{i}$ by the corresponding $\taubar_{\rho(i)}$,
we can rewrite the objective function in terms of $\taubar$ as $\left\langle \bar{\theta},\taubar\right\rangle $
where the coefficients $\bar{\theta}$ are defined on nodes and edges
of the lifted graph $\bar{\gm}$ as follows. For each node orbit $\vbar$,
$\bar{\theta}_{\vbar:t}=\sum_{v'\in\vbar}\oparaunary{v'}t=|\vbar|\oparaunary vt$
where $t\in\{0,1\}$ and $v$ is any representative of $\vbar$. For
each edge orbit $\bar{e}$ with a representative $\{u,v\}\in\bar{e}$,
$\bar{\theta}_{\bar{e}:tt}=\sum_{\{u',v'\}\in\bar{e}}\oparabinary{u'}{v'}tt=|\bar{e}|\oparabinary uvtt$
where $t\in\{0,1\}$, $\bar{\theta}_{(\overline{u,v}):01}=\sum_{(u',v')\in(\overline{u,v})}\oparabinary{u'}{v'}01=|(\overline{u,v})|\oparabinary uv01$.
Note that typically the two arc-orbits $(\overline{u,v})$ and $(\overline{v,u})$
are not the same, in which case $|(\overline{u,v})|=|(\overline{v,u})|=|\bar{e}|$.
However, in case $(\overline{u,v})=(\overline{v,u})$ then $|(\overline{u,v})|=|(\overline{v,u})|=2|\bar{e}|$.

So, we have shown that the lifted formulation for MAP inference on
the local polytope can be described in terms of the lifted variables
$\taubar$ and the lifted parameters $\bar{\theta}$. These lifted
variables and parameters are associated with the orbits of the ground
graphical model. Thus, the derived lifted formulation can also be
read out directly from the lifted graph $\bar{\gm}$. In fact, the
derived lifted formulation is the local relaxed MAP problem of the
lifted graphical model $\bar{\gm}$. Therefore, any algorithm for
solving the local relaxed MAP problem on $\gm$ can also be used to
solve the derived lifted formulation on $\bar{\gm}$. From lifted
inference point of view, we can lift any algorithm for solving the
local relaxed MAP problem on $\gm$ by constructing $\bar{\gm}$ and
run the same algorithm on $\bar{\gm}$. This allows us to lift even
asynchronous message passing algorithms such as the max-product linear
programming (MPLP) algorithm~\citep{globerson07fixing}, which cannot
be lifted using existing lifting techniques.\textbf{}%

\section{Beyond Local Polytope: Lifted MAP Inference with Cycle Inequalities}

We now discuss lifting the MAP relaxation on $\Cyc(\gm)$, a bound
obtained by tightening $\Local(\gm)$ with an additional set of linear
constraints that hold on cycles of the graphical model structure $\gm$,
called cycle constraints~\citep{sontag07newouter}. These constraints
arise from the fact that the number of cuts (transitions from 0 to
1 or vice versa) in any configuration on a cycle of $\gm$ must be
even. Cycle constraints can be framed as linear constraints on the
mean vector $\omean$ as follows. For every cycle $C$ (set of edges
that form a cycle in $\gm$) and every odd-sized subset $F\subseteq C$
\begin{equation}
\sum_{\{u,v\}\in F}nocut(\{u,v\},\tau)+\sum_{\{u,v\}\in C\backslash F}cut(\{u,v\},\tau)\ge1\label{eq:cycle-constraint}
\end{equation}
where $nocut(\{u,v\},\tau)=\pmeanbinary uv00+\pmeanbinary uv11$ and
$cut(\{u,v\},\tau)=\pmeanbinary uv01+\pmeanbinary vu01$.

Theorem \ref{theorem:lifted-outer} guarantees that MAP inference
on $\Cyc$ can be lifted by restricted the feasible domain to $\Cyc_{\shatter^{o}}$,
which we term the \emph{lifted cycle polytope}. Substituting the original
variables $\tau$ by the lifted variables $\taubar$, we obtain the
\emph{lifted cycle constraints} in terms of $\taubar$
\begin{equation}
\sum_{\{u,v\}\in F}nocut(\{\overline{u,v}\},\taubar)+\sum_{\{u,v\}\in C\backslash F}cut(\{\overline{u,v}\},\taubar)\ge1\label{eq:cycle-constraint-lifted}
\end{equation}
where $nocut(\{\overline{u,v}\},\taubar)=\liftpmeanbinary{\{\overline{u,v}\}}00+\liftpmeanbinary{\{\overline{u,v}\}}11$
and $cut(\{\overline{u,v}\},\taubar)=\liftpmeanbinary{(\overline{u,v})}01+\liftpmeanbinary{(\overline{v,u})}01$
where $(\overline{u,v})$ and $(\overline{v,u})$ are the arc-orbits
corresponding to the node-orbit $\{\overline{u,v}\}$.

\subsection{Lifted Cycle Constraints on All Cycles Passing Through a Fixed Node}

\global\long\def\cyc{\text{Cyc}}

Fix a node $i$ in $\gm$, and let $\cyc[i]$ be the set of cycle
constraints generated from all cycles passing through $i$. A cycle
is simple if it does not intersect with itself or contain repeated
edges; \citep{sontag07newouter} considers only simple cycles, but
we will also consider any cycle, including non-simple cycles in $\cyc[i]$.
Adding non-simple cycles to the mix does not change the story since
constraints on non-simple cycles of $\gm$ are redundant. We now give
a precise characterization of $\overline{\cyc}[i]$, the set of lifted
cycle constraints obtained by lifting all cycle constraints in $\cyc[i]$
via the transformation from (\ref{eq:cycle-constraint}) to (\ref{eq:cycle-constraint-lifted}).

The lifted graph fixing $i$, $\bar{\gm}[i]$ is defined as follows.
Let $\Aut_{\parti}[\FF,i]$ be the subgroup of $\Aut_{\parti}[\FF]$
that fixes $i$, that is $\naut(i)=i$. The set of nodes of $\bar{\gm}[i]$
is the set of node orbits $\bar{V}[i]$ of $\gm$ induced by $\Aut_{\parti}[\FF,i]$,
and the set of edges is the set of edge orbits $\bar{E}[i]$ of $\gm$.
Each edge orbit connects to the orbits of the two adjacent nodes (which
could form just one node orbit). Since $i$ is fixed, $\{i\}$ is
a node orbit, and hence is a node on $\bar{\gm}[i]$. Note that $\bar{\gm}[i]$
in general is not a simple graph: it can have multi-edges and loops. 
\begin{thm}
\label{theorem:lifted-cycle-through-one}Let $\bar{C}$ be a cycle
(not necessarily simple) in $\bar{\gm}[i]$ that passes through the
node $\{i\}$. For any odd-sized $\bar{F}\subset\bar{C}$
\begin{equation}
\sum_{\ebold\in\bar{F}}nocut(\ebold,\taubar)+\sum_{\ebold\in\bar{C}\backslash\bar{F}}cut(\ebold,\taubar)\ge1\label{eq:lifted-cycle-through-i}
\end{equation}
is a constraint in $\overline{\cyc}[i]$. Furthermore, all constraints
in $\overline{\cyc}[i]$ can be expressed this way.
\end{thm}

\subsection{Separation of lifted cycle constraints}

While the number of cycle constraints may be reduced significantly
in the lifted space, it may still be computationally expensive to
list all of them. To address this issue, we follow \citep{sontag07newouter}
and employ a cutting plane approach in which we find and add only
the most violated lifted cycle constraint in each iteration (separation
operation). 

For finding the most violated lifted cycle constraint, we propose
a lifted version of the method presented by \citep{sontag07newouter},
which performs the separation by iterating over the nodes of the graph
$\gm$ and for each node $i$ finds the most violated cycle constraint
from all cycles passing through $i$. %
{} Theorem 7.1 suggests that all lifted cycle constraints in $\overline{\cyc}[i]$
can be separated by mirroring $\bar{\gm}[i]$ and performing a shortest
path search from $\{i\}$ to its mirrored node, similar to the way
separation is performed on ground cycle constraints \citep{sontag07newouter}.

To find the most violated lifted cycle constraint, we could first
find the most violated lifted cycle constraint $C_{i}$ in $\overline{\cyc}[i]$
for each node $i$, and then take the most violated constraints over
all $C_{i}$. However, note that if $i$ and $i'$ are in the same
node orbit, then $\overline{\cyc}[i]=\overline{\cyc}[i']$. Hence,
we can perform separation using the following algorithm:
\begin{enumerate}
\item For each node orbit $\vbar\in\bar{V}$, choose a representative $i\in\vbar$
and find its most violated lifted cycle constraint $C_{\vbar}\in\overline{\cyc}[i]$
using a shortest path algorithm on the mirror graph of $\bar{\gm}[i]$. 
\item Return the most violated constraint over all $C_{\vbar}$. 
\end{enumerate}
Notice that both $\bar{\gm}[i]$ and its mirror graph have to be calculated
only once per graph. In each separation iteration we can reuse these
structures, provided that we adapt the edge weights in the mirror
graph according to the current marginals.

\section{\label{sec:Detecting-symmetries}Detecting Symmetries in Exponential
Families}

\subsection{\label{sub:Sym-Nauty}Detecting Symmetries via Graph Automorphisms}

We now discuss the computation of a subgroup of the automorphism group
$\Aut_{\parti}(\FF)$. Our approach is to construct a suitable graph
whose automorphism group is guaranteed to be a subgroup of $\Aut_{\parti}(\FF)$,
and thus any tool and algorithm for computing graph automorphism can
be applied. The constructed graph resembles a factor graph representation
of $\FF$. However, we also use colors of factor nodes to mark feature
functions that are identical and in the same cell of $\parti$, and
colors of edges to encode symmetry of the feature functions themselves.
\begin{defn}
The colored factor graph induced by $\FF$ and $\parti$, denoted
by $\gr_{\parti}[\FF]$ is a bipartite graph with nodes $V(\gr)=\left\{ x_{1}\ldots x_{n}\right\} \cup\left\{ \nrfeat_{i}\ldots\nrfeat_{m}\right\} $
and edges $E(\gr)=\left\{ \left\{ x_{\argn_{i}(k)},f_{i}\right\} \,|\, i\in\Index,\: k=1\ldots\argnall i\right\} $.
Variable nodes are assigned the same color which is different from
the colors of factor nodes. Factor nodes $\nrfeat_{i}$ and $\nrfeat_{j}$
have the same color iff $\nrfeat_{i}\equiv\nrfeat_{j}$ and $i\simparti{\parti}j$.
If the function $\nrfeat_{i}$ is symmetric, then all edges adjacent
to $\nrfeat_{i}$ have the same color; otherwise, they are colored
according to the argument number of $f_{i}$, i.e., $\left\{ x_{\argn_{i}(k)},\nrfeat_{i}\right\} $
is assigned the $k$-th color.%
\end{defn}
\begin{thm}
\label{theorem:colored-graph-aut}The automorphism group \textup{$\Aut[\mathfrak{G}_{\parti}]$
of $\gr_{\parti}[\FF]$} is a subgroup of \textup{$\Aut_{\parti}(\FF)$},
i.e., $\Aut[\mathfrak{G}_{\parti}]\subgroup\Aut_{\parti}[\FF]$. 
\end{thm}
Finding the automorphism group $\Aut[\mathfrak{G}_{\parti}]$ of the
graph $\gr_{\parti}[\FF]$ therefore yields a procedure to compute
a subgroup of $\Aut_{\parti}[\FF]$. Thus, according to corollary
\ref{cor:lifting-partition-mean}, the induced orbit partition on
the factor node of $\gr_{\parti}[\FF]$ is a lifting partition for
the variational problems discussed earlier. \emph{Nauty}, for example,
directly supports the operation of computing the automorphism group
of a graph and extracting the induced node orbits.

\subsection{\label{sub:Sym-MLN}Symmetries of Markov Logic Networks}

A Markov Logic Network (MLN) \citep{richardson06markov} is prescribed
by a list of weighted formulas $\formula_{1}\ldots\formula_{K}$ (consisting
of a set of predicates, logical variables, constants, and a weight
vector ${\bf w}$) and a logical domain $\ldom=\{a_{1}...a_{|\ldom|}\}$.
Let $\lconsts$ be the set of objects appearing as constants in these
formulas, then $\ldomnovel=\ldom\backslash\lconsts$ is the set of
objects in $\ldom$ that do not appear in these formulas. Let $\Ground$
be the set of all ground predicates $p(a_{1}\ldots a_{\ell})$'s.
If $s$ is a substitution, $\formula_{i}[s]$ denotes the result of
applying the substitution $s$ to $F_{i}$ and is a grounding of $\formula_{i}$
if it does not contain any logical free variables. The set of all
groundings of $\formula_{i}$ is $\GroundFormula_{i}$, and let $\GroundFormula=\GroundFormula_{1}\cup\ldots\cup\GroundFormula_{K}$.
The MLN corresponds to an exponential family $\FF_{MLN}$ where $\Ground$
is the variable index set and each grounding $\formula_{i}[s]\in\GroundFormula_{i}$
is a feature function $\feat_{\formula_{i}[s]}(\model)=\mathbb{I}(\model\vDash\formula_{i}[s])$
with the associated parameter $\theta_{F_{i}[s]}=w_{i}$ where $\model$
is a truth assignment to all the ground predicates in $\Ground$ and
$w_{i}$ is the weight of the formula $\formula_{i}$. Since all the
ground features of the formula $\formula_{i}$ have the same parameter
$w_{i}$, the MLN also induces the parameter-tying partition $\parti_{MLN}=\{\{\feat_{\formula_{1}[s]}(\model)\}\dots\{\feat_{\formula_{K}[s]}(\model)\}\}$. 

Let a renaming permutation $\rename$ be a permutation over $\ldom$
that fixes every object in $\lconsts$, i.e., $\rename$ only permutes
objects in $\ldomnovel$$ $. Thus, the set of all such renaming permutations
is a group $\mathbb{G}^{re}$ that is isomorphic to the symmetric
group $\symg(\ldomnovel)$. Consider the following actions of $\Gg^{re}$
on $\Ground$ and $\GroundFormula$: $\naut_{\rename}:\; p(a_{1}\ldots a_{\ell})\mapsto p(\rename(a_{1})\ldots\rename(a_{\ell}))$
and $\paut_{\rename}:\formula_{i}[s]\mapsto\formula_{i}[\rename(s)]$
where $\rename(s=(x_{1}/a_{1},...,x_{k}/a_{k}))=(x_{1}/\rename(a_{1}),...,x_{k}/\rename(a_{k}))$.
Basically, $\naut_{\rename}$ and $\paut_{\rename}$ rename the constants
in each ground predicate $p(a_{1}\ldots a_{\ell})$ and ground formula
$\formula_{i}[s]$ according to the renaming permutation $\rename$.
The following theorem (a consequence of Lemma 1 from \citet{bui12lifted})
shows that $\Gg^{re}$ is isomorphic to a subgroup of $\Aut[\FF_{MLN}]$,
the automorphism group of the exponential family $\FF_{MLN}$.
\begin{thm}
\label{theorem:renaming-aut}For every renaming permutation $r$,
$(\naut_{\rename},\paut_{\rename})\in\Aut[\FF_{MLN}]$. Thus, $\Gg^{re}\preceq\Aut[\FF_{MLN}]$. 
\end{thm}
Furthermore, observe that $\paut_{\rename}$ only maps between groundings
of a formula $\formula_{i}$, thus the action of $\Gg^{re}$ on $\GroundFormula$
is consistent with the parameter-tying partition $\parti_{MLN}=\{\{\feat_{\formula_{1}[s]}(\model)\}\dots\{\feat_{\formula_{K}[s]}(\model)\}\}$.
Thus, $\Gg^{re}\preceq\Aut_{\parti_{MLN}}[\FF_{MLN}]$. According
to corollary \ref{cor:lifting-partition-mean}, the orbit partition
induced by the action of $\Gg^{re}$ on $\GroundFormula$ is a lifting
partition for the variational inference problems associated with the
exponential family $\FF_{MLN}$. In addition, this orbit partition
can be quickly derived from the first-order representation of an MLN;
the size of this orbit partition depends only on the number of observed
constants $|\ldom_{o}|$, and does not depend on actual domain size
$|\ldom|$.

\section{Experiments}

We experiment with several propositional and lifted methods for variational
MAP inference by varying the domain size of the following MLN:
\begin{eqnarray*}
w_{1} & x\neq y\wedge x\neq z\wedge y\neq z\Rightarrow\text{pred}\left(x,y\right)\Leftrightarrow\text{pred}\left(y,z\right)\\
w_{2} & \text{x\ensuremath{\neq}y}\wedge\text{obs}\left(x,y\right)\Rightarrow\text{pred}\left(x,y\right)\\
 & \text{obs}(A,B)
\end{eqnarray*}
This MLN is designed to be a simplified version of models that enforce
transitivity for the predicate $\text{pred}$, and will be called
the \emph{semi-transitive} model.%
\footnote{If $pred(x,y)=1$ is interpreted as having a (directed) edge from
$x$ to $y$, then this model represents a random graph whose nodes
are elements of the domain of the MLN. More specifically, the model
can be thought of as a 2-star Markov graph \cite{frank86markovgraph}.%
} %
{} We set the weights as $w_{1}=-100$ and $w_{2}=0.1$. The negative
$w_{1}$ yields a repulsive model with relatively strong interaction,
while the shared predicate and variables in the first formula are
known to be a difficult case for lifted inference. The third formula
is an observation with two constants $A$ and $B$. 

The ground Markov network of the above MLN is corresponding to an
exponential family $\FF_{MLN}$, and we use the two methods described
in Sections \ref{sub:Sym-Nauty} and \ref{sub:Sym-MLN} to derive
lifting partitions. The first method (\emph{nauty}) fully grounds
the MLN, then finds a lifting partition using nauty. The second (\emph{renaming})
works directly with the MLN, and uses the renaming group to find a
lifting partition. We use two outer bounds to the marginal polytope:
$\Local$ and $\Cyc$. There are three variants of each method: propositional,
lifting using nauty orbit partition, and lifting using renaming orbit
partition. This yields a total of six methods to compare. For reference,
we also calculate the exact solution to the MAP problem using ILP. 

Figure \ref{fig:dom-runtime} shows the runtime (in milliseconds)
until convergence for different domain sizes of the logical variables
in our MLN. We can make a few observations. First, in most cases lifting
dramatically reduces runtime for larger domains. Second, nauty-based
methods suffer from larger domain sizes. This is expected, as we perform
automorphism finding on propositional graphs with increasing size.
Third, the renaming partition outperforms nauty partitions, by virtue
of working directly with the first-order representation. Notice in
particular for lifted-via-renaming methods, we can still observe a
dependency on domain size, but this is an artifact of our current
implementation---in the future these curves will be constant. Finally,
all but the propositional cycle method are faster than ILP. 

Figure \ref{fig:obj} illustrates how the objective changes over cutting
plane iterations (and hence time), all for the case of domain size
10. Both the local polytope and ILP approaches have no cutting plane
iterations, and hence are represented as single points. Given that
ILP is exact, the ILP point gives the optimal solution. Notice how
all methods are based on outer/upper bounds on the variational objective,
and hence are decreasing over time. First, we can observe that the
$\Cyc$ methods converge to the (almost) optimal solution, substantially
better than the $\Local$ methods. However, in the propositional case
the $\Cyc$ algorithm converges very slowly, and is only barely faster
than ILP. 

Lifted CYCLE methods are the clear winners for this problem. We can
also see how the different lifting partitions affect $\Cyc$ performance.
The renaming partition performs its first iteration much quicker than
the nauty-based partition, since nauty needs to work on the full grounded
network. Consequently, it converges much earlier, too. However, we
can also observe that the renaming partition is more fine-grained
than the nauty partition, leading to larger orbit graphs and hence
slower iterations. Notably, working with lifted cycle constraints
gives us substantial runtime improvements, and effectively optimal
solutions. 

\begin{center}
\begin{figure}[t]
\begin{centering}
\subfloat[\label{fig:dom-runtime}Runtime vs. domain size. ]{\begin{centering}
\includegraphics[scale=0.35]{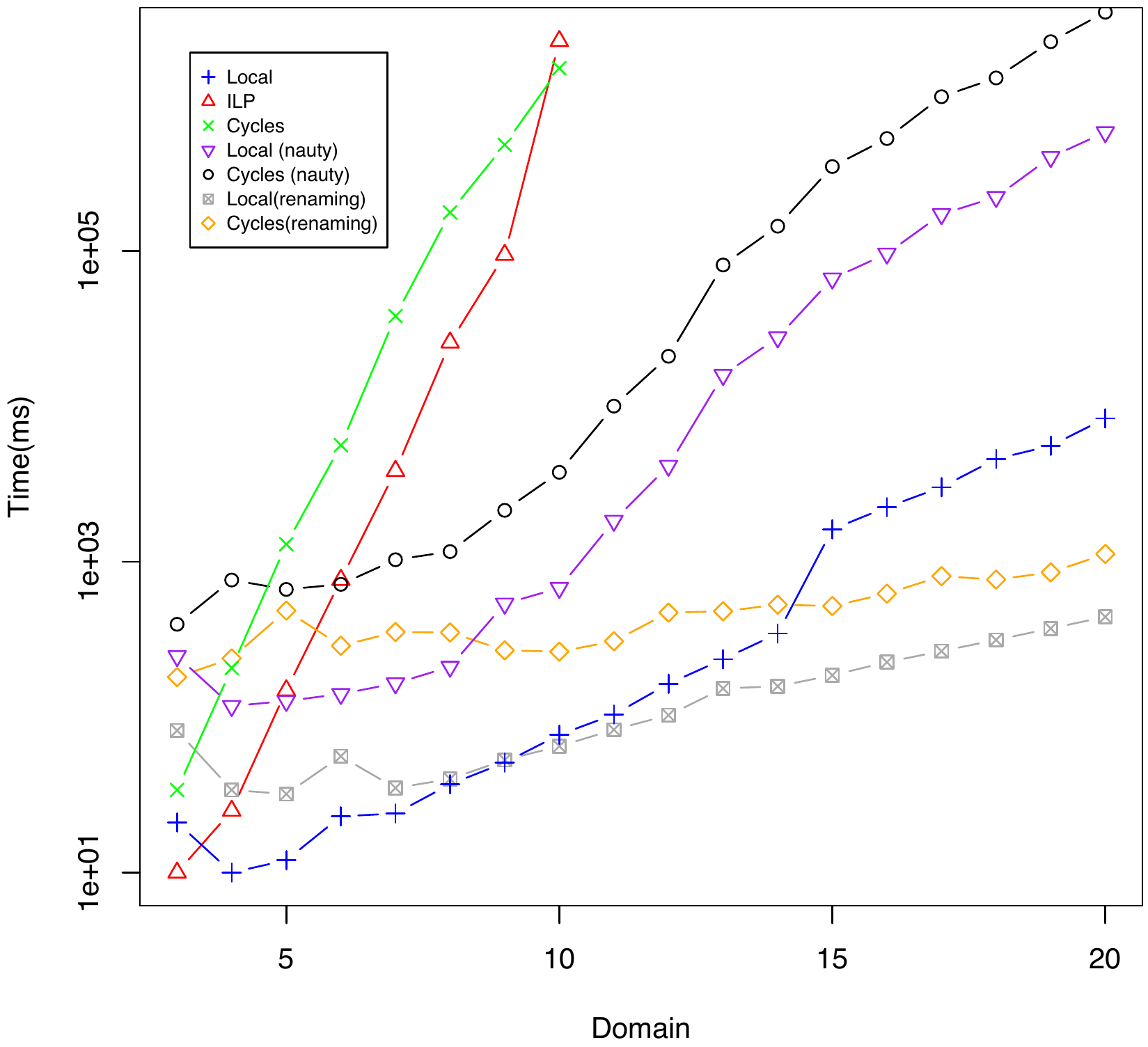}
\par\end{centering}

}\vspace{-3mm}
\par\end{centering}

\centering{}\subfloat[\label{fig:obj}Objective over time for domain size 10.]{\begin{centering}
\includegraphics[scale=0.35]{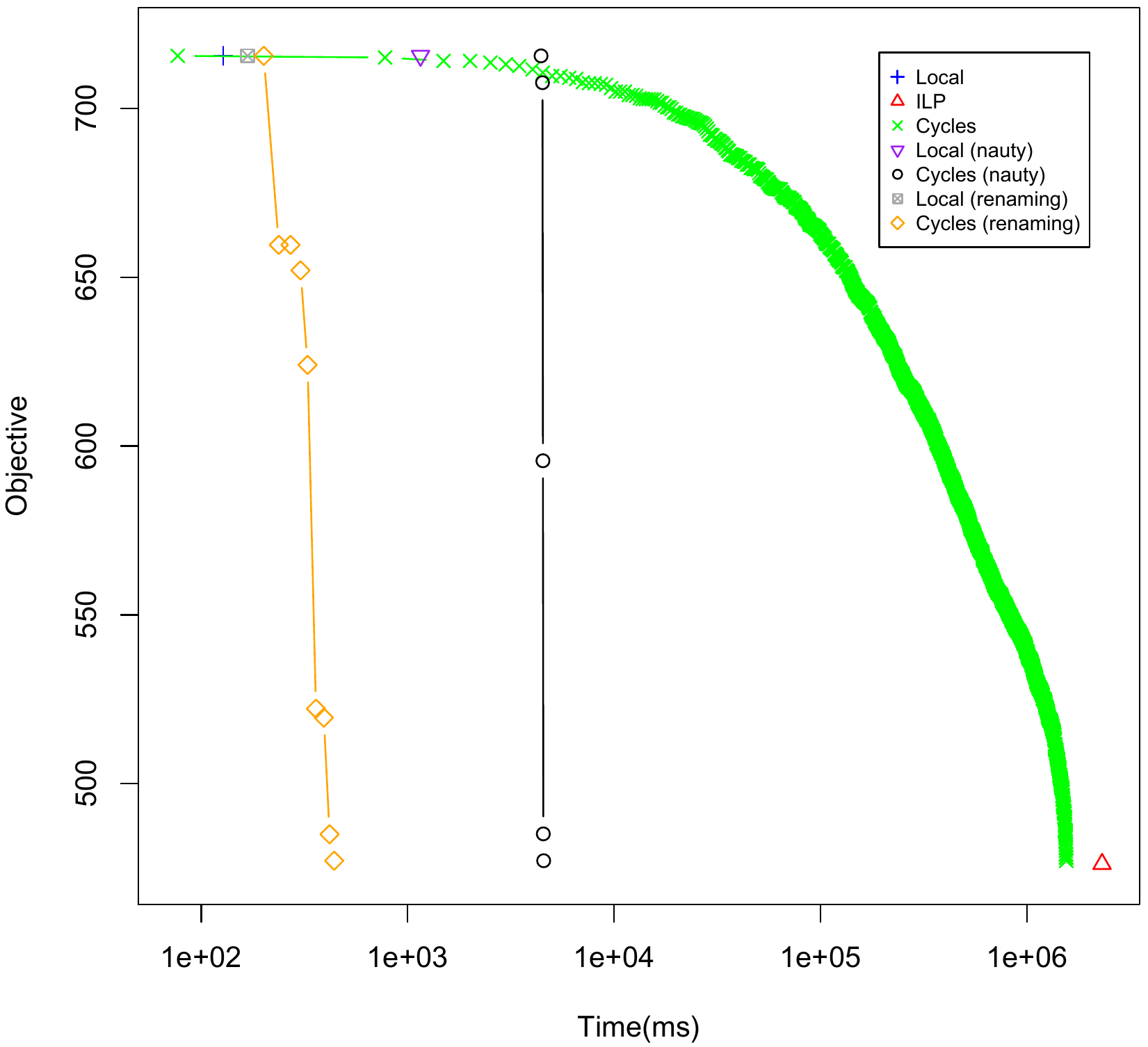}
\par\end{centering}

}\caption{\label{fig:Experiments}Experiments with a semi-transitive model with
one observed variable. Due to large differences between runtimes,
time is always presented in logarithmic scale.}
\end{figure}

\par\end{center}

\section{Conclusion}

We presented a new general framework for lifted variational inference.
In doing this, we introduce and study a precise mathematical definition
of symmetry of graphical models via the construction of their automorphism
groups. Using the device of automorphism groups, orbits of random
variables are obtained, and lifted variational inference is materialized
as performing the corresponding convex variational optimization problem
in the space of per-orbit random variables. Our framework enables
lifting a large class of approximate variational MAP inference algorithms,
including the first lifted algorithm for MAP inference with cycle
constraints. We presented experimental results demonstrating that
lifted MAP inference with cycle constraints achieved the state of
the art performance, obtaining much better objective function values
than LOCAL approximation while remaining relatively efficient. Our
future work includes extending this approach to handle approximations
of convex upper-bounds of $A^{*}$, which would enable lifting the
full class of approximate convex variational marginal inference.

\section{Proofs}

\textbf{Proof of proposition \ref{prop:aut-characterization}.}
\begin{proof}
(Part 1) We first prove that if $(\naut,\paut)\in\Aut[\FF]$ then
the conditions in the theorem hold. Pick $i\in\Index$ and let $\paut(i)=j$.
Since $\Feat^{\paut}(x)=\Feat(x^{\naut})$, $\feat_{j}(x)=\feat_{i}(x^{\naut})$.
Express the feature $\feat_{i}$ and $\feat_{j}$ in their factorized
forms, we have $\nrfeat_{j}(x_{j_{1}}\ldots x_{j_{\argnall j}})=\nrfeat_{i}(x_{\naut(i_{1})}\dots x_{\naut(i_{\argnall i})})$.
Since $\nrfeat_{j}$ cannot be reduced further, it must depend on
all the distinct arguments in $\{j_{1}\ldots j_{\argnall j}\}$. This
implies that the set of arguments on the LHS $\{\naut(i_{1})\ldots\naut(i_{\argnall i})\}\supset\{j_{1}\ldots j_{\argnall j}\}$.
Thus $\argnall i\ge\argnall j$. Apply the same argument with the
automorphism $(\naut^{-1},\paut^{-1})$, and note that $\paut^{-1}(j)=i$,
we obtain $\argnall j\ge\argnall i$. Thus $\argnall i=\argnall j=K$.
Furthermore, $\{\naut(i_{1})\ldots\naut(i_{K})\}=\{j_{1}\ldots j_{K}\}$.
This implies that $\naut$ is a bijection from $\args(\nrfeat_{i})=\{i_{1}\ldots i_{K}\}$
to $\args(\nrfeat_{j})=\{j_{1}\ldots j_{K}\}$.

For the third condition, from $\nrfeat_{j}(x_{j_{1}}\ldots x_{j_{\argnall j}})=\nrfeat_{i}(x_{\naut(i_{1})}\dots x_{\naut(i_{\argnall i})})$,
we let $t_{k}=x_{j_{k}}$ so that $t_{\argn_{j}^{-1}(k)}=x_{k}$ (since
$j_{k}=\argn_{j}(k)$) to arrive at $\nrfeat_{j}(t_{1}\ldots t_{K})=\nrfeat_{i}(t_{\argn_{j}^{-1}\circ\naut\circ\argn_{i}(1)}\ldots t_{\argn_{j}^{-1}\circ\naut\circ\argn_{i}(K)})$,
or in short form $\nrfeat_{j}(t)=\nrfeat_{i}(t^{\alpha})$. $\alpha$
is a bijection since all the mappings $\argn_{j}$, $\argn_{i}$ and
$\naut$ are bijections.

(Part 2) Let $(\naut,\paut)$ be a pair of permutations such that
the three conditions are satisfied, we will show that they form an
automorphism of $\FF$. Pick $i\in\Index$ and let $j=\paut(i)$ and
$K=\argnall i=\argnall j$. From $\nrfeat_{j}(t)=\nrfeat_{i}(t^{\alpha})$,
we have $\nrfeat_{j}(x_{j_{1}}\ldots x_{j_{K}})=\nrfeat_{i}(x_{j_{\alpha(1)}}\ldots x_{j_{\alpha(K)}})$.
Note that $j_{\alpha(k)}=\argn_{j}\circ\alpha(k)=\naut(i_{k})$. Thus
$\nrfeat_{j}(x_{j_{1}}\ldots x_{j_{K}})=\nrfeat_{i}(x_{\naut(i_{1})}\ldots x_{\naut(i_{K})})$,
so $\feat_{j}(x)=\feat_{i}(x^{\naut})$ and hence $\Feat^{\paut}(x)=\Feat(x^{\naut})$. 
\end{proof}
\textbf{Proof of theorem \ref{theorem:aut-properties}.} 
\begin{proof}
Part (1) To prove that $\naut$ is an automorphism of $\gm$, the
hypergraph representing the structure of the exponential family graphical
model, we need to show that $c\subset\vIndex$ is a hyperedge (cluster)
of $\gm$ iff $\naut(c)$ is a hyperedge. 

If $c$ is a hyperedge, $\exists i\in\Index$ such that $c=\args(\nrfeat_{i})$.
Let $j=\paut(i)$, by proposition \ref{prop:aut-characterization},\textbf{
$\naut(c)=\args(\nrfeat_{j})$}, so $\naut(c)$ is also a hyperedge. 

If $\naut(c)$ is an hyperedge, apply the same reasoning using the
automorphism $(\naut^{-1},\paut^{-1})$, we obtain $\naut^{-1}(\naut(c))=c$
is also a hyperedge.

Part (2)-(5) We first state some identities that will be used repeatedly
throughout the proof. Let $x,y\in\Real^{n}$. The first identity states
that permuting two vectors do not change their inner products
\begin{equation}
\left\langle x,y\right\rangle =\left\langle x^{\naut},y^{\naut}\right\rangle \label{eq:permute-inner-prod}
\end{equation}
As a result if $(\naut,\paut)\in\Aut[\FF]$ 
\begin{equation}
\left\langle \Feat(x^{\naut}),\para^{\paut}\right\rangle =\mbox{\ensuremath{\left\langle \Feat^{\paut^{-1}}(x^{\naut}),\para\right\rangle }}=\left\langle \Feat(x),\para\right\rangle \label{eq:permute-feature-inner-prod}
\end{equation}
The next identity allows us to permute the integrating variable in
a Lebesgue integration 
\begin{equation}
\int_{S}f(x)d\lambda=\int_{S^{\naut^{-1}}}f(x^{\naut})d\lambda\label{eq:permute-integral}
\end{equation}
where $\lambda$ is a counting measure, or a Lebesgue measure over
$\Real^{n}$. The case of counting measure can be verified directly
by establishing a bijection between summands of the two summations,
and the case of Lebesgue measure is direct result of the property
of linearly transformed Lebesgue integrals (Theorem 24.32, page 616
\cite{yeh2006realanalysis}).

Part (2). By definition of the log-partition function, 
\begin{eqnarray*}
A(\para^{\paut}) & = & \int_{\domain^{n}}h(x)\exp\left\langle \Feat(x),\para^{\paut}\right\rangle d\lambda\\
 & = & \int_{\domain^{n}}h(x^{\naut})\exp\left\langle \Feat(x^{\aut}),\para^{\paut}\right\rangle d\lambda\text{ (by \ref{eq:permute-integral})}\\
 & = & \int_{\domain^{n}}h(x)\exp\left\langle \Feat(x),\para)\right\rangle d\lambda\text{ (by \ref{eq:permute-feature-inner-prod}})\\
 & = & A(\para)
\end{eqnarray*}

As a result, $\Para^{\paut}=\left\{ \para^{\paut}|A(\para)<\infty\right\} =\left\{ \para^{\paut}|A(\para^{\paut})<\infty\right\} =\Para$.

Part (3). $\FF(x^{\naut}|\para^{\paut})=h(x^{\naut})\exp\left\langle \Feat(x^{\naut}),\para^{\paut}\right\rangle =h(x)\exp\left\langle \Feat(x),\para\right\rangle =\FF(x|\para)$.

Part (4). Expand $\meanmap^{\paut}(\para)$ gives
\[
\meanmap^{\paut}(\para)=\Expt_{\para}\Feat^{\paut}(x)=\int_{\domain^{n}}\Feat^{\paut}(x)\FF(x|\para)d\lambda=\int_{\domain^{n}}\Feat^{\paut}(x^{\naut^{-1}})\FF(x^{\naut^{-1}}|\para)d\lambda
\]
 where the last equality follows from (\ref{eq:permute-integral}).
Since $(\naut^{-1},\paut^{-1})$ is also an automorphism, $\Feat(x^{\naut^{-1}})=\Feat^{\paut^{-1}}(x)$,
thus $\Feat^{\paut}(x^{\naut^{-1}})=\Feat(x)$. Further, by part (3),
$\FF(x^{\naut^{-1}}|\para)=\FF(x|\para^{\paut})$. Thus $\meanmap^{\paut}(\para)=\int_{\domain^{n}}\Feat(x)\FF(x|\para^{\paut})=\meanmap(\para^{\paut})$.

Part (5). Let $\mean\in\Mean$, so $\mean=\int_{\domain^{n}}p(x)\Feat(x)d\lambda$
for some probability density $p$. Expand $\mean^{\paut}$ gives 
\[
\mean^{\paut}=\int_{\domain^{n}}p(x)\Feat^{\paut}(x)d\lambda=\int_{\domain^{n}}p(x)\Feat(x^{\naut})d\lambda=\int_{\domain^{n}}p(x^{\naut^{-1}})\Feat(x)d\lambda
\]
Let $p'(x)=p(x^{\naut^{-1}})$ and observe that $\int p'(x)d\lambda=\int p(x)d\lambda=1$,
so $p'$ is also a probability density. Thus $\mean^{\paut}\in\Mean$,
hence $\Mean^{\paut}\subset\Mean$. Apply similar reasoning to the
automorphism $(\naut^{-1},\paut^{-1})$, we have $\mean^{\paut^{-1}}\in\Mean.$
Thus, every $\mean\in\Mean$ can be expressed as $\mean'^{\paut}$
for some $\mean'\in\Mean$, but this means $\Mean\subset\Mean^{\paut}$.
Thus, $\Mean=\Mean^{\paut}$.

For $\mean\in\relint\Mean$, there exists $\para\in\Para$ such that
$\mean=\meanmap(\para)$. The negative entropy function becomes $A^{*}(\mean)=\left\langle \mean,\para\right\rangle -A(\para)$.
From part (4), $\mean^{\paut}=\meanmap(\para^{\paut})$, thus $A^{*}(\mean^{\paut})=\left\langle \mean^{\paut},\para^{\paut}\right\rangle -A(\para^{\paut})=\left\langle \mean,\para\right\rangle -A(\para)=A^{*}(\mean)$.

For $\mean\in\text{border }\Mean\backslash\relint\Mean$, $A^{*}(\mean^{\paut})=A^{*}(\mean)$
holds by a continuity argument.
\end{proof}
\textbf{Proof of theorem \ref{theorem:lifting-convex-optimization}.}
\begin{proof}
The proof makes use of the orbit-stabilizer theorem, an elementary
group-theoretic result which we describe below. 

Let $\Gg$ be a finite group acting on $\Index$, and $i\in\Index$.
Let $\orbit(i)=\left\{ k|\exists g\in\Gg\text{ s.t. }g(i)=k\right\} $
be the orbit containing $i$ and let $\stab(i)=\left\{ g\in\Gg|g(i)=i\right\} $
be the stabilizer of $i$. The orbit-stabilizer theorem essentially
states that the group $\Gg$ can be partitioned into $|\orbit(i)|$
subsets $\Gg=\cup_{k\in\orbit(i)}G_{k}$ where $G_{k}=\left\{ g\in\Gg|g(i)=k\right\} $
for each $k\in\orbit(i)$, and $|G_{k}|=|\stab(i)|$. Thus $|\Gg|=|\orbit(i)||\stab(i)|$.

As a consequence, we can simplify summation over group elements to
an orbit sum

\begin{equation}
\frac{1}{|\Gg|}\sum_{g\in\Gg}f(g(i))=\frac{1}{|\Gg|}\sum_{k\in\orbit(i)}\sum_{g\in G_{k}}f(g(i))=\frac{1}{|\orbit(i)|}\sum_{k\in\orbit(i)}f(k)\label{eq:group-orbit-sum}
\end{equation}

We now return to the main proof of the theorem. Note that $\inf_{S}J(x)=c$
is equivalent to $\forall x\in S,\ J(x)\ge c$ and there exists a
sequence $\{x_{(n)}\}\subset S$ such that $J(x_{(n)})\rightarrow c$
($c$ can be $-\infty$). Clearly, $J(x)\ge c$ $\forall x\in S_{\shatter}$,
so all we need to establish is a sequence $\{x_{(n)}^{*}\}\subset S_{\shatter}$
such that $J(x_{(n)}^{*})\rightarrow c$.

Let $x\in S\subset\Real^{m}$. Since $\Gg$ stabilizes $S$, $x^{g}\in S$
for all $g\in\Gg$. Define $x^{*}=\frac{1}{|\Gg|}\sum_{g\in\Gg}x^{g}$
as the \emph{symmetrization} of $x$. Since $S$ is convex, $x^{*}\in S$.
Since $J$ is convex and $\Gg$ stabilizes $J$, $J(x^{*})\le\frac{1}{|\Gg|}\sum_{g\in\Gg}J(x^{g})=J(x)$.

Consider one element of $x_{i}^{*}$ of the vector $x^{*}$. Using
(\ref{eq:group-orbit-sum}), we can express $x_{i}^{*}$ as the average
of $x_{k}$ for all $k$ in $i$'s orbit 
\begin{equation}
x_{i}^{*}=\frac{1}{\Gg}\sum_{g\in\Gg}x_{g(i)}=\frac{1}{|\orbit(i)|}\sum_{k\in\orbit(i)}x_{k}\label{eq:symmetrized-vector}
\end{equation}
so if $i$ and $j$ are in the same orbit, $x_{i}^{*}=x_{j}^{*}$.
Thus, $x^{*}\in S_{\shatter}$. 

With the above construction, we obtain a sequence $\{x_{(n)}^{*}\}\subset S_{\shatter}$
such that $c\le J(x_{(n)}^{*})\le J(x_{(n)})$. Since $J(x_{(n)})\rightarrow c$,
we also have $J(x_{(n)}^{*})\rightarrow c$. Thus, $\inf_{x\in S_{\shatter}}J(x)=c$.
\end{proof}
\textbf{Proof of corollary \ref{cor:lifting-partition-mean}.}
\begin{proof}
Observe that the group $\Aut_{\parti}[\FF]$ stabilizes the set $\Mean$,
the function $A^{*}(\mean)$ (theorem \ref{theorem:aut-properties},
part (6)) and the linear function $\left\langle \para,\mean\right\rangle $
when the coefficient $\para\in\Para_{\parti}$. Thus, this result
is a direct consequence of theorem \ref{theorem:lifting-convex-optimization}. 
\end{proof}
\textbf{Proof of theorem \ref{theorem:lifted-extremes}}.
\begin{proof}
Recall that if $g\in\Aut_{\parti}[\FF]$ then $g=(\naut,\paut)$.
The group $\Aut_{\parti}[\FF]$ acts on $\Index$ by the permuting
action of $\paut$ and on $\vIndex$ by the permuting action of $\naut$.
We thus write $x^{g}$ to denote $x^{\naut}$, and $\Feat^{g}(x)$
to denote $\Feat^{\paut}(x)$.

Consider the symmetrization of $\Feat(x)$, defined as $\Feat^{*}(x)=\frac{1}{|\Aut_{\parti}[\FF]|}\sum_{g\in\Aut_{\parti}[\FF]}\Feat^{g}(x)$.
Using an argument similar to (\ref{eq:symmetrized-vector}), $\Feat^{*}(x)\in\Real_{\shatter}^{m}$.
Clearly, $\Feat^{*}(x)\in\Mean$, so $\Feat^{*}(x)\in\Mean_{\shatter}$.
One the other hand, since $g\in\Aut[\FF]$, $\Feat^{g}(x)=\Feat(x^{g})$,
so 
\[
\Feat^{*}(x)=\frac{1}{|\Aut_{\parti}[\FF]|}\sum_{g\in\Aut_{\parti}[\FF]}\Feat(x^{g})=\frac{1}{|\mathcal{C}(x)|}\sum_{y\in\mathcal{C}(x)}\Feat(y)=\bar{\Feat}(\mathcal{C}(x))
\]
 where we have used (\ref{eq:group-orbit-sum}) and $\mathcal{C}(x)=\orbit_{\Aut_{\parti}[\FF]}(x)$
is the orbit containing $x$.

We now return to the main proof. From the above, we have $\bar{\Feat}(\mathcal{C})\in\Mean_{\shatter}$,
so clearly $\chull\left\{ \bar{\Feat}(\mathcal{C})|\mathcal{C}\in\mathcal{O}\right\} \subset\Mean_{\shatter}$.
Now, let $\mean\in\Mean_{\shatter}$, then $\mean=\sum_{x\in\domain^{n}}p(x)\Feat(x)$
for some probability distribution $p$. Furthermore, $\mean^{g}=\mean$
for all $g\in\Aut_{\parti}[\FF]$. Thus
\[
\mean=\frac{1}{|\Aut_{\parti}[\FF]|}\sum_{g\in\Aut_{\parti}[\FF]}\mean^{g}=\frac{1}{|\Aut_{\parti}[\FF]|}\sum_{g\in\Aut_{\parti}[\FF]}\sum_{x\in\domain^{n}}p(x)\Feat^{g}(x)=\sum_{x\in\domain^{n}}p(x)\bar{\Feat}(\mathcal{C}(x))=\sum_{\mathcal{C}\in\mathcal{O}}p(\mathcal{C})\bar{\Feat}(\mathcal{C})
\]
where $p(\mathcal{C})=\sum_{y\in\mathcal{C}}p(y)$. Therefore, $\mean\in\chull\left\{ \bar{\Feat}(\mathcal{C})|\mathcal{C}\in\mathcal{O}\right\} $,
so $\Mean_{\shatter}\subset\chull\left\{ \bar{\Feat}(\mathcal{C})|\mathcal{C}\in\mathcal{O}\right\} $.
\end{proof}
\textbf{Proof of corollary \ref{cor:lifted-variational-overcomplete}.}
\begin{proof}
Let $\gm=\gm[\FF]$. If $\naut$ is an automorphism of $\gm$ then
$\pi$ induces a permutation on $\oIndex$ which we denoted by $\naut^{o}$.
We proceed in two steps. Step (1): if $\naut\in\Aut[\gm]$ then $(\naut,\naut^{o})\in\Aut[\FF^{o}]$
where $\FF^{o}$ is the overcomplete family induced from $\FF$; this
guarantees that $\Aut_{\parti}[\FF]$, via the action $\naut^{o}$
stabilizes $\oMean$ and $A^{o^{*}}$. Step (2): if $(\naut,\paut)\in\Aut_{\parti}[\FF]$
and $\para\in\Para_{\parti}$ then $(\opara)^{\naut^{o}}=\opara$;
this guarantees that $\Aut_{\parti}[\FF]$ stabilizes the linear function
$\left\langle \opara,\omean\right\rangle $, again via the action
$\naut^{o}$. These two steps together with theorem \ref{theorem:lifting-convex-optimization}
will complete the proof.

Step (1). Recall that $\naut^{o}(u:t)=\naut(u):t$ and $\naut^{o}(\left\{ u:t,v:t'\right\} )=\left\{ \naut(u):t,\naut(v):t'\right\} $.
Note that $\naut^{o}$ is well-defined only if $\naut$ is an automorphism
of $\gm$. We will show that $\Feat^{o}(x^{\naut})=(\Feat^{o}(x))^{\naut^{o}}$.
Indeed 
\[
\ofeatunary ut(x^{\naut})=\idf{x_{\naut(u)}=t}=\ofeatunary{\naut(u)}t(x)
\]
\[
\ofeatbinary uvt{t'}(x^{\naut})=\idf{x_{\naut(u)}=t,x_{\naut(v)}=t'}=\ofeatbinary{\naut(u)}{\naut(v)}t{t'}(x)
\]

Step(2). Note that if $(\naut,\paut)\in\Aut[\FF]$ then $\paut$ is
a bijection between $\left\{ i|\args(\feat_{i})=S\right\} $ and $\left\{ j|\args(\feat_{j})=\naut(S)\right\} $.
Furthermore, if $(\naut,\paut)\in\Aut_{\parti}[\FF]$ then $\para_{\paut(i)}=\para_{i}$
for all $i\in\Index$. 

For $u\in\vIndex$ 
\begin{eqnarray*}
\opara_{u:t} & = & \sum_{i\ \args(\feat_{i})=\{u\}}\nrfeat_{i}(t)\para_{i}=\sum_{i\ \args(\feat_{i})=\{u\}}\nrfeat_{\paut(i)}(t)\para_{\paut(i)}\\
 & = & \sum_{j\ \args(\feat_{j})=\{\naut(u)\}}\nrfeat_{j}(t)\para_{j}=\opara_{\naut(u):t}
\end{eqnarray*}
where $\nrfeat_{i}(t)=\nrfeat_{\paut(i)}(t)$ follows from proposition
\ref{prop:aut-characterization}.

For $\left\{ u,v\right\} \in E(\gm)$, without loss of generality,
assume $u<v$. Take $i\in\Index$ such that $\args(\feat_{i})=\left\{ u,v\right\} $.
By proposition \ref{prop:aut-characterization}, if $\naut(u)<\naut(v)$
then $\nrfeat_{i}(t,t')=\nrfeat_{\paut(i)}(t,t')$ and
\begin{eqnarray*}
\oparabinary uvt{t'} & = & \sum_{i\ \args(\feat_{i})=\{u,v\}}\nrfeat_{i}(t,t')\para_{i}=\sum_{i\ \args(\feat_{i})=\{u,v\}}\nrfeat_{\paut(i)}(t,t')\para_{\paut(i)}\\
 & = & \sum_{j\ \args(\feat_{j})=\{\naut(u),\naut(v)\}}\nrfeat_{j}(t,t')\para_{j}=\opara_{\{\naut(u):t,\naut(v):t'\}}
\end{eqnarray*}

If $\naut(u)>\naut(v)$ then by proposition \ref{prop:aut-characterization}
$\nrfeat_{i}(t,t')=\nrfeat_{\paut(i)}(t',t)$ and
\begin{eqnarray*}
\oparabinary uvt{t'} & = & \sum_{i\ \args(\feat_{i})=\{u,v\}}\nrfeat_{i}(t,t')\para_{i}=\sum_{i\ \args(\feat_{i})=\{u,v\}}\nrfeat_{\paut(i)}(t',t)\para_{\paut(i)}\\
 & = & \sum_{j\ \args(\feat_{j})=\{\naut(u),\naut(v)\}}\nrfeat_{j}(t',t)\para_{j}=\opara_{\{\naut(u):t,\naut(v):t'\}}
\end{eqnarray*}

\end{proof}
\textbf{Proof of theorem \ref{theorem:lifted-outer}.}
\begin{proof}
From the proof of corollary \ref{cor:lifted-variational-overcomplete},
$\Aut_{\parti}[\FF]$ stabilizes the objective function $\left\langle \opara,\omean\right\rangle $,
so it remains to show that this group also stabilizes the set $\outbound$.

We first elaborate on what it means in a formal sense for $\outbound$
to depend only on the graph $\gm$. The intuition here is that the
constraints that form $\outbound$ are constructed purely from graph
property of $\gm$, and not from the way we assign label to nodes
of $\gm$. Formally, let $I_{\outbound}(\pmean,\gm)$ be the indicator
function of the set $\outbound$: given a pair $(\pmean,\gm)$, this
function return $1$ if $\pmean$ belongs to $\outbound(\gm)$ and
$0$ otherwise. Relabeling $\gm$ by assigning the index $\naut(u)$
to the node $u$ for some $\naut\in\symg_{n}$, we obtain a graph
$\gm'=\gm^{\naut}$ isomorphic to $\gm$. Reassign the index of $\pmean$
accordingly, we obtain $\pmean^{(\naut^{o})}$. Since construction
of $\outbound$ is invariant w.r.t. relabeling of $\gm$, we have
$I_{\outbound}(\pmean,\gm)=I_{\outbound}(\pmean^{\naut^{o}},\gm^{\naut})$. 

If $\naut$ is an automorphism of $\gm$, $I_{\outbound}(\pmean,\gm)=I_{\outbound}(\pmean^{\naut^{o}},\gm)$,
so $\tau\in\outbound(\gm)\Leftrightarrow\pmean^{\naut^{o}}\in\outbound(\gm)$.
Thus the group $\Aut(\gm)$ stabilizes $\outbound(\gm)$. From theorem
\ref{theorem:aut-properties}, if $(\naut,\paut)\in\Aut[\FF]$ then
$\naut$ is an automorphism of $\gm$. Thus, $\Aut_{\parti}[\FF]$
also stabilizes $\outbound(\gm)$.
\end{proof}
\textbf{Proof of theorem \ref{theorem:lifted-cycle-through-one}.}
\begin{proof}
Clearly every lifted cycle constraint in $\overline{\cyc}[i]$ can
be rewritten in form (\ref{eq:lifted-cycle-through-i}). We now show
that every constraint in this form is a lifted constraint in $\overline{\cyc}[i]$.
To do this, for every cycle $\bar{C}$ passing through $\{i\}$ and
every odd-sized $\bar{F}\subset\bar{C}$, we will point out a constraint
in $\cyc[i]$ whose lifted form is of the form (\ref{eq:lifted-cycle-through-i}).

We first show that if ${\bf e}$ is an edge orbit connecting two node
orbits ${\bf u}$ and ${\bf v}$, then for any $u\in{\bf u}$, there
exists an edge $e=\{u,v\}$ such that $e\in{\bf e}$ and $v\in{\bf v}$.
Let $\{u_{o},v_{o}\}$ be an arbitrary member of ${\bf e}$ such that
$u_{0}\in{\bf u}$ and $v_{0}\in{\bf v}$. Since $u$ and $u_{0}$
are in the same node orbit, there exists a group element $g$ such
that $g(u_{0})=u$. Take $v=g(v_{0})$, then clearly $e=\{u,v\}$
satisfies $e\in{\bf e}$ and $v\in{\bf v}$. 

Using the above, it is straight forward to prove a stronger statement
by induction. If ${\bf p}={\bf e}_{1},\ldots,{\bf e}_{n}$ is a path
in $\bar{\gm}[i]$ from node orbit ${\bf u}$ to ${\bf v}$, and let
$u\in{\bf u}$, then there exists a path $p=e_{1},\ldots,e_{n}$ in
$\gm$ from node $u$ to $v$ such that $e_{j}\in{\bf e}_{j}$ for
all $j$, and $v\in{\bf v}$. 

A cycle in $\bar{\gm}[i]$ passing through $\{i\}$ is a path $\bar{C}={\bf e}_{1},\ldots,{\bf e}_{n}$
from $\{i\}$ to $\{i\}$ itself. Thus, there must exist a path $C=e_{1},\ldots,e_{n}$
in $\gm$ from $i$ to $i$ (so that $C$ is a cycle in $\gm$ passing
through $i$), and $e_{j}\in{\bf e}_{j}$. Thus, take an arbitrary
constraint of the form (\ref{eq:lifted-cycle-through-i}), there exists
a corresponding ground constraint on the cycle $C$ passing through
$i$ in $\gm$, and this constraint clearly belongs to $\cyc[i]$. 
\end{proof}
\textbf{Proof of theorem \ref{theorem:colored-graph-aut}.}
\begin{proof}
Since $\gr_{\parti}$ is a bi-partite graph and variable and factor
nodes have different colors, an automorphism of $\gr_{\parti}$ must
have a form of a pair of permutation $(\naut,\paut)$ where $\naut\in\symg_{n}$
is a permutation among variable nodes and $\paut\in\symg_{m}$ is
a permutation among factor nodes. 

Let $j=\paut(i)$. Since $i$ and $j$ have the same color, $j\simparti{\parti}i$.
This shows that $\paut$ is consistent with the partition $\parti$.

We now show that $(\naut,\paut)$ is an automorphism of the exponential
family $\FF$. To do this, we make use of proposition \ref{prop:aut-characterization}.
From the coloring of $\gr_{\parti}$ we have $\nrfeat_{i}\equiv\nrfeat_{j}$.
Since $\naut$ maps neighbors of $i$ to neighbors of $j$, $\naut$
must be a bijection from $\args(\nrfeat_{i})$ to $\args(\nrfeat_{j})$.
Let $\alpha=\argn_{j}^{-1}\circ\naut\circ\argn_{i}$, we need to show
that $\nrfeat_{i}(t^{\alpha})=\nrfeat_{j}(t)$ for all $t$. There
are two cases. 

(i) If $\nrfeat_{i}$ is a symmetric function, so is $\nrfeat_{j}$
and thus $\nrfeat_{i}(t^{\alpha})=\nrfeat_{i}(t)=\nrfeat_{j}(t)$.

(ii) If $\nrfeat_{i}$ is not a symmetric function, since $\naut$
must preserve the coloring of edges adjacent to $i$ and $j$, it
must map $\nrfeat_{i}$'s $k$-th argument to $\nrfeat_{j}$'s $k$-th
argument: $\naut(\argn_{i}(k))=\argn_{j}(k)$. Therefore $\alpha(k)=\argn_{j}^{-1}(\argn_{j}(k))=k$,
so $\alpha$ is the identity permutation. Thus, $\nrfeat_{i}(t^{\alpha})=\nrfeat_{i}(t)=\nrfeat_{j}(t)$.
\end{proof}
\textbf{Proof of theorem \ref{theorem:renaming-aut}.}
\begin{proof}
Let $r$ be a renaming permutation, and let $\omega$ be a Herbrand
model. Let $r(\omega)$ denote the Herbrand model obtained by applying
$r$ to all groundings in $\omega$. Using lemma 1 from \cite{bui12lifted},
we have $\omega\models F_{k}(s)$ iff $r(\omega)\models F_{k}(r(s))$.
Writing $\omega$ as a vector of $0$ or $1$, where $1$ indicates
that the corresponding grounding is true, then $r(\omega)$ in vector
form is the same as $\omega^{\aut_{r}^{-1}}$, e.g., the vector $\omega$
permuted by $\naut_{r}^{-1}$. Thus, $\idf{\omega\models F_{k}(s)}=\idf{\omega^{\naut_{r}^{-1}}\models\paut_{r}(F_{k}(s))}$,
or equivalently, if $\Feat$ is the feature function of the MLN in
vector form, then $\Feat(\omega)=\Feat^{\paut_{r}}(\omega^{\naut_{r}^{-1}})$.
Thus $(\naut_{r},\paut_{r})$ is an automorphism of the MLN. \end{proof}

\begin{acknowledgement*}
The authors gratefully acknowledge the support of the Defense Advanced
Research Projects Agency (DARPA) Machine Reading Program under Air
Force Research Laboratory (AFRL) prime contract no. FA8750-09-C-0181.
Any opinions, findings, and conclusions or recommendations expressed
in this material are those of the author(s) and do not necessarily
reflect the view of DARPA, AFRL, or the U.S. government.

Distribution Statement \textquotedblleft{}A\textquotedblright{} (Approved
for Public Release, Distribution Unlimited).
\end{acknowledgement*}
\bibliographystyle{plainnat}
\bibliography{riedel}

\end{document}